\documentclass[accepted]{uai2023} 

\usepackage[british]{babel}

\usepackage{natbib} 
\usepackage{mathtools} 
\usepackage{booktabs} 
\usepackage{tikz} 

\usepackage{cleveref}
\usepackage{microtype}
\usepackage{graphicx}
\usepackage{subcaption}
\usepackage{bbm}
\usepackage{amsfonts}
\usepackage{amsmath} 
\usepackage{mathtools}
\newtheorem{definition}{Definition}
\usepackage{multirow}
\usepackage{siunitx}
\DeclareMathOperator*{\argmin}{arg\,min}

\usepackage{xr}   
\usepackage{hyperref}

\title{Quantifying lottery tickets under label noise:\\ accuracy, calibration, and complexity}

\author[1]{\href{mailto:<varora@sissa.it>?Subject=Your UAI 2023 paper}{Viplove Arora}{}}
\author[2]{Daniele Irto}
\author[1]{Sebastian Goldt}
\author[1]{Guido Sanguinetti}
\affil[1]{%
    Theoretical and Scientific Data Science\\
    SISSA\\
    Trieste, Italy
}
\affil[2]{%
    Data Science and Scientific Computing\\
    University of Trieste\\
    Trieste, Italy
}

\makeatletter

\begin{document}
\maketitle

\begin{abstract}
Pruning deep neural networks is a widely used strategy to alleviate the computational burden in machine learning. Overwhelming empirical evidence suggests that pruned models retain very high accuracy even with a tiny fraction of parameters. However, relatively little work has gone into characterising the small pruned networks obtained, beyond a measure of their accuracy. In this paper, we use the sparse double descent approach to identify univocally and characterise pruned models associated with classification tasks. We observe empirically that, for a given task, iterative magnitude pruning (IMP) tends to converge to networks of comparable sizes even when starting from full networks with sizes ranging over orders of magnitude. We analyse the best pruned models in a controlled experimental setup and show that their number of parameters reflects task difficulty and that they are much better than full networks at capturing the true conditional probability distribution of the labels. On real data, we similarly observe that pruned models are less prone to overconfident predictions. Our results suggest that pruned models obtained via IMP not only have advantageous computational properties but also provide a better representation of uncertainty in learning. 
\end{abstract}

\section{Introduction}
\label{sec:intro}
Conventional statistical wisdom suggests that increasing the size of a model leads to an initial improvement in generalisation performance, followed by dramatic overfitting and degradation of accuracy in highly overparameterised models. In reality, the implicit regularisation of large models leads to a double descent which, under suitable conditions, leads to overparameterised models with even stronger generalisation performance~\citep{vallet1989linear, opper1990ability, geman1992neural, belkin2019reconciling, nakkiran2021deep, loog2020brief}. This phenomenon is well understood theoretically in simple cases and has been replicated in practice in a variety of modern deep neural networks architectures~\citep{belkin2019reconciling, nakkiran2021deep, arpit_generalization, neyshabur_generalization, belkin2020two}. 

\begin{figure*}
    \centering
    \includegraphics[width=.49\linewidth]{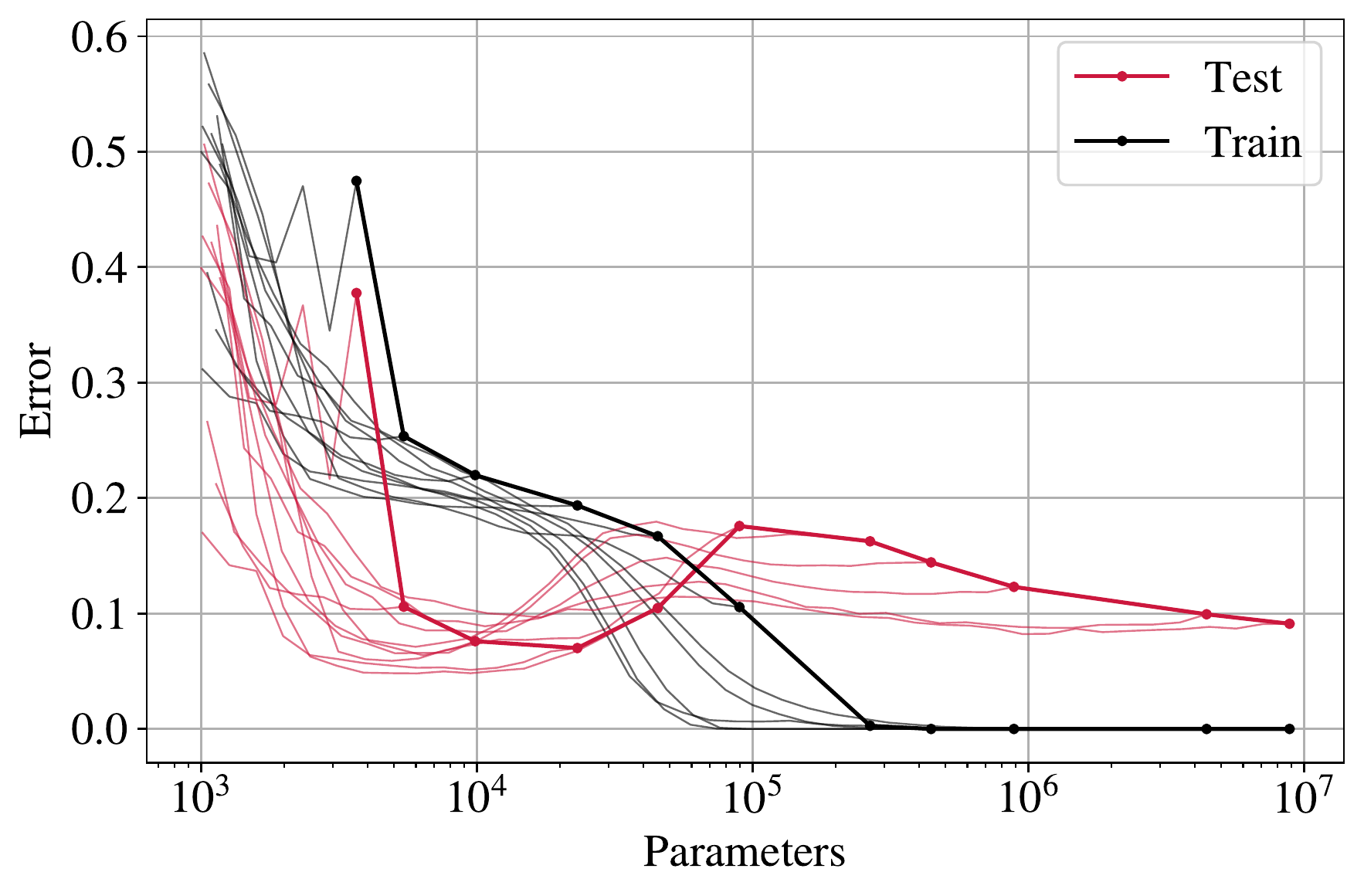} \hfill
    \includegraphics[width=.49\linewidth]{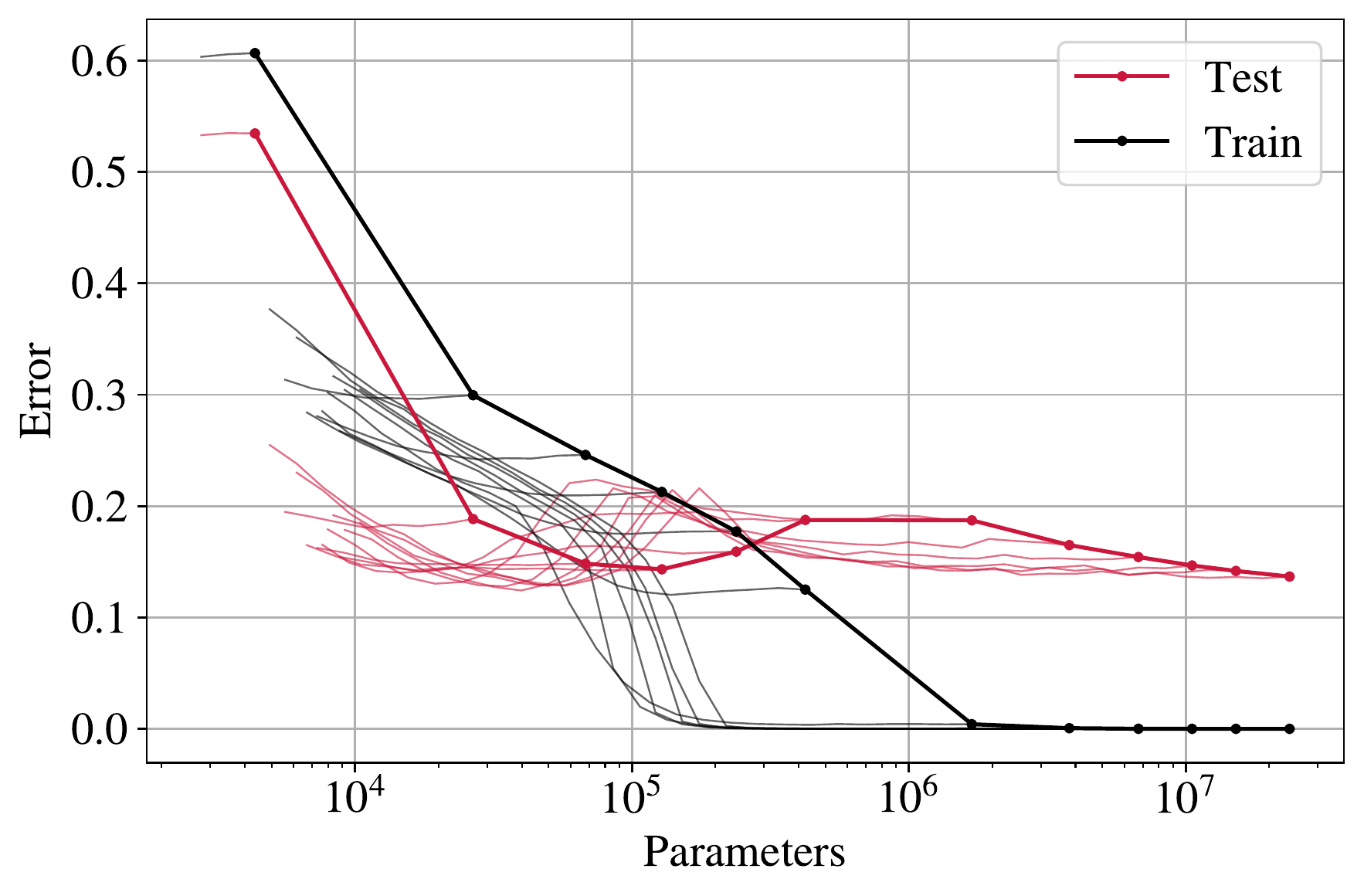}
    \caption{Pruning models along the double descent curve (dark red) shows that sparse double descent curves (light red) from different models coincide at the minima. Results are shown for three-layer FCNs on MNIST (left) and ResNet-18 on CIFAR-10 (right) with 20\% label noise averaged over three replicates.}
    \label{fig:sdd_n20}
\end{figure*}

However, decades of research on pruning neural networks \citep{lecun1989optimal, hassibi1992second, gale2019state, hoefler2021sparsity} has also shown that a large number of parameters (weights) in these overparameterized models can be removed without compromising on the generalisation error. \citet{he2022sparse} recently reported that iterative pruning leads to a converse phenomenon to double descent, a {\it sparse double descent}: generalisation performance initially degrades upon pruning, and then improves to reach an optimum frequently providing even better performance than the original full model. The sparse double descent allows us to identify an optimal model by plotting the generalisation error as a function of the number of parameters during the iterative pruning process. How the architectural bias induced by pruning enables the networks to reverse the double descent curve is not understood theoretically, and is relatively unexplored empirically.

Here we seek to address this knowledge gap in the iterative magnitude pruning (IMP) framework \citep{han2015learning}, a simple and successful approach for pruning deep neural networks \citep{blalock2020state, renda2020comparing} which is pivotal to finding ``lottery tickets''~\citep{frankle2018lottery, frankle2019stabilizing, frankle2020linear}, i.e.~sparse subnetworks that can be trained in isolation without compromising on accuracy. We start with a remarkable empirical observation. \Cref{fig:sdd_n20} shows the double descent curve (thick line) and the sparse double descent curves (thin lines) for fully connected networks trained on MNIST (left) and a ResNet-18 trained on CIFAR-10 (right). Naturally, the starting point of the double descent curve is fixed at the smallest possible model. \citet{he2022sparse} considered sparse double descent starting from a single, large model. Here, we consider sparse double descent curves starting from neural networks with a large range of parameters. We show in \cref{fig:sdd_n20} that irrespective of the initial full model size (or indeed test accuracy), all sparse double descent curves tend to achieve their minimum (best pruned model) within a very narrow range of model sizes, which is systematically smaller than the trough of the first descent (best small full model). What is special about such sparse models? What enables the improvement in generalisation by pruning?

To answer this question quantitatively, we investigate the behaviour of the best pruned models in a controlled binary classification scenario where data is generated from distributions of differing complexity. We find that the size of the best pruned model correlates well with task complexity. Moreover, best pruned models are considerably better at capturing the underlying true conditional label distribution than either full models (which tend to be significantly overconfident) or best small full models. These empirical observations are replicated in our analysis of real data, suggesting that IMP indeed captures important aspects of the uncertainty in the learning problem, as well as producing computationally more tractable models. 

We see a key contribution of our empirical study in suggesting a way to estimate the number of ``effective'' parameters that a trained model contains, and that are required for solving a classification task. Quantifying the effective number of parameters in trained neural networks has been a central question for the theory of neural networks for a long time, see for example~\citet{breiman1995reflections}, yet it remains an open challenge. In this manuscript, we focus on the empirical findings and leave the theoretical analysis for future work. Our main contributions are:

\begin{enumerate}
    \item We find that IMP prunes models of different sizes to produce sparse double descent curves that coincide at the minima of the test error. We define the \textit{effective number of parameters} required for a given classification task and model/architecture. We demonstrate this phenomenon in fully-connected and convolutional neural networks trained on MNIST and CIFAR-10. Additionally, the effective number of parameters is comparable across architectures, suggesting that our procedure identifies an intrinsic property of the classification task.
    \item By training neural networks on binary Gaussian mixture classification tasks of increasing difficulty, we show that the effective number of parameters in a model correlates with task difficulty.
    \item We finally study the calibration of pruned models and show that pruned models capture the true distribution of the synthetic data models better than their unpruned counterparts. As a consequence, we show that the best pruned models are better at capturing uncertainty in their predictions.
\end{enumerate}

\section{Related Work}
\label{sec:related_work}
Dating back to 1998, \citet{poppi1998optimal} used pruning to find the optimal neural network architecture. They found that pruning could improve generalisation and even recover the optimal model in a linear dataset. \citet{kuhn2021robustness} introduces a new complexity measure for neural networks, namely the fraction of weights that can be pruned from the network without affecting its performance. They found that the fraction of prunable weights increases with network width for a ResNet trained on CIFAR-10. \citet{li2018measuring} uses the idea of subspace training to estimate the number of parameters (or intrinsic dimension) needed to achieve good performance using neural networks. They found that many problems have smaller intrinsic dimensions than one might suspect, and the intrinsic dimension for a given dataset varies little across a family of models with vastly different sizes. While these conclusions align with our findings, their results rely on the usage of random subspace solutions with a performance $\approx 90\%$ of the baseline to define the intrinsic dimension. Our approach instead uses sparsification to find the effective number of parameters using models that generalise better than the baseline. 

\citet{venkatesh2020calibrate} used different calibration strategies on overparameterised models to study the impact on the resulting lottery tickets. \citet{leicalibrating} propose a new sparse training method that improves the reliability of pruned models. From a theoretical perspective, \citet{sakamoto2022analyzing} used PAC-Bayesian theory to understand the generalisation behaviour of pruned networks. \citet{zhang2021lottery} characterise the performance of training a pruned neural network to show that pruning enlarges the convex region near a desirable model with guaranteed generalisation. \citet{yang2023theoretical} used a controlled setting under random pruning to determine pruning fractions that can improve generalisation performance. A series of theoretical works \citep{malach2020proving, orseau2020logarithmic, pensia2020optimal, da2022proving, burkholz2022most} have shown the existence of a winning ticket inside larger (deeper and wider) networks of different sizes, thus providing some intuition on the amount of overparameterisation required. 

To understand how IMP can improve the generalisation of neural networks by acting as a regulariser, \citet{jin2022pruning} studied the loss of influential samples in the optimally pruned models. {\citet{paul2022unmasking} use the geometry of the error landscape at each level of pruning to understand the principles behind the success of IMP (with rewinding) without label noise. Our focus is primarily on the properties (accuracy, number of parameters, and calibration) of the best pruned models.} \citet{ankner2022effect} used the framework of \citet{pope2021intrinsic}, which uses GANs to generate images with known intrinsic dimensions, to find that the intrinsic dimensionality of data correlates with the prunability of neural networks. Certain efforts have also been made to understand the masks learned using pruning \citep{paganini2020iterative, pellegrini2021sifting}.

Since the rediscovery of the double descent behaviour in deep neural networks~\citep{belkin2019reconciling, nakkiran2021deep}, characterising double descent curves from simple models to deep networks has become a very active area of research, see \citet{hastie2019surprises, spigler2019jamming, advani2020highdimensional,  belkin2019reconciling, dascoli2020double,  lin2020causes, dascoli2020triple, mei2022generalization} for a small sample. Here, our focus is not on this double descent phenomenon -- instead, our goal is to provide a univocal procedure to associate a pruned model with a large model using the sparse double descent.

\section{Experimental Setup}
\label{sec:setup}
\paragraph{Datasets:} We used MNIST~\citep{lecun1998gradient} and CIFAR-10~\citep{krizhevsky2009learning} for our experiments with real data. Additional experiments were also performed using the Fashion-MNIST dataset~\citep{xiao2017fashion} (see \cref{fig:fmnist} for results). We also perform experiments in a controlled setting, where we consider binary mixture classification with inputs sampled from a Gaussian mixture in $D=100$ dimensions. This simple setting for the data model allows us to precisely compute the true conditional class probability of data and characterise the actual decision boundary. This allows us to closely investigate the architectural bias induced by pruning. We consider two settings for mixture classification. The \emph{linear} dataset consists of two clusters that can be separated using a linear classifier. The two clusters have means $\mu_1 \neq \mu_2$, but same covariance matrix $\Sigma_1 = \Sigma_2$. The \emph{XOR} dataset was created such that the resulting Gaussians (that have the same covariance) are placed like the graphic representation of the XOR logical function. The training and test sets contain $\num{10000}$ and $\num{5000}$ samples respectively. See \cref{app:mixt_class} for more details. 

A crucial aspect of our experiments is adding symmetric label noise {by randomly permuting the labels for a fraction of} the training data. Adding label noise to training data provides a straightforward way to produce double descent \citep{nakkiran2021deep} and sparse double descent \citep{he2022sparse} in deep neural networks. {Note that using other kinds of label noise could provide different conclusions, but is not the focus of the current study and hence is out of scope for this paper.} While adding label noise seems unrealistic, \citet{northcutt2021pervasive} found that labelling errors are pervasive in several benchmark machine learning datasets and lower capacity models may be more practical. Building on this observation, we show that pruning can potentially provide a way of finding this capacity. 

\paragraph{Models:} A two-layer fully-connected network (FCN) was used for our experiments on the Gaussian mixture classification tasks with five replicates for each model (see details in \cref{app:hyperparameters}). The width of the hidden layer was varied to obtain models of varying sizes and produce the double descent curve. For MNIST, we used two and three-layer FCNs while varying the width of the first hidden layer, and ResNet-6 with convolutional filters of different widths. For CIFAR-10, we used ResNet-18 \citep{he2016deep} and varied the width of the convolutional filter to obtain the double descent curve. Each experiment was replicated three times for real-world datasets. The hyperparameters used in these experiments are described in \cref{app:hyperparameters}. The cross-entropy loss function was used to train all the models. For CIFAR-10, we also successively removed one class at a time and then trained and pruned a ResNet-18 model with a fixed width of the convolutional filter. All our models were trained long enough to ensure that they were not in the epoch-wise double descent regime \citep{nakkiran2021deep}. The \texttt{OpenLTH} library\footnote{\scriptsize{\url{https://github.com/facebookresearch/open_lth}}} was used for our experimental evaluations. The code for replicating our experiments is available on GitHub\footnote{\scriptsize{\url{https://github.com/viplovearora/noisy_lottery_tickets}}}.

\paragraph{Network pruning:} IMP with 20\% weights removal at each iteration (lottery ticket rewinding was used when required, see \cref{app:IMP}) was used for our experiments as it provides an effective procedure to find subnetworks with nontrivial sparsities that have low test error \citep{frankle2018lottery, frankle2020linear, blalock2020state, renda2020comparing}. It should be noted that other pruning techniques like random pruning and gradient-based pruning also produce the sparse double descent curve \citep{he2022sparse} and can be used instead.

\begin{figure*}[!ht]
\begin{subfigure}{\textwidth}
\centering
    \begin{subfigure}{.19\textwidth}
    \centering
        \includegraphics[width=\linewidth]{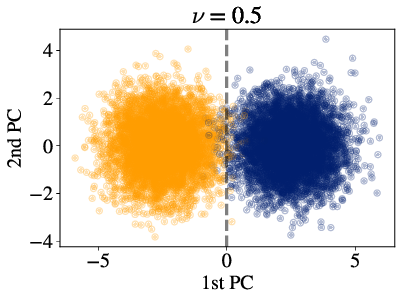}\\
        \includegraphics[width=\linewidth]{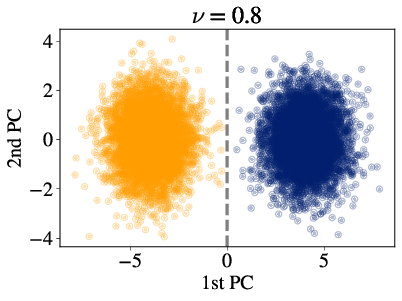}
    \end{subfigure}
    \includegraphics[width=.4\linewidth]{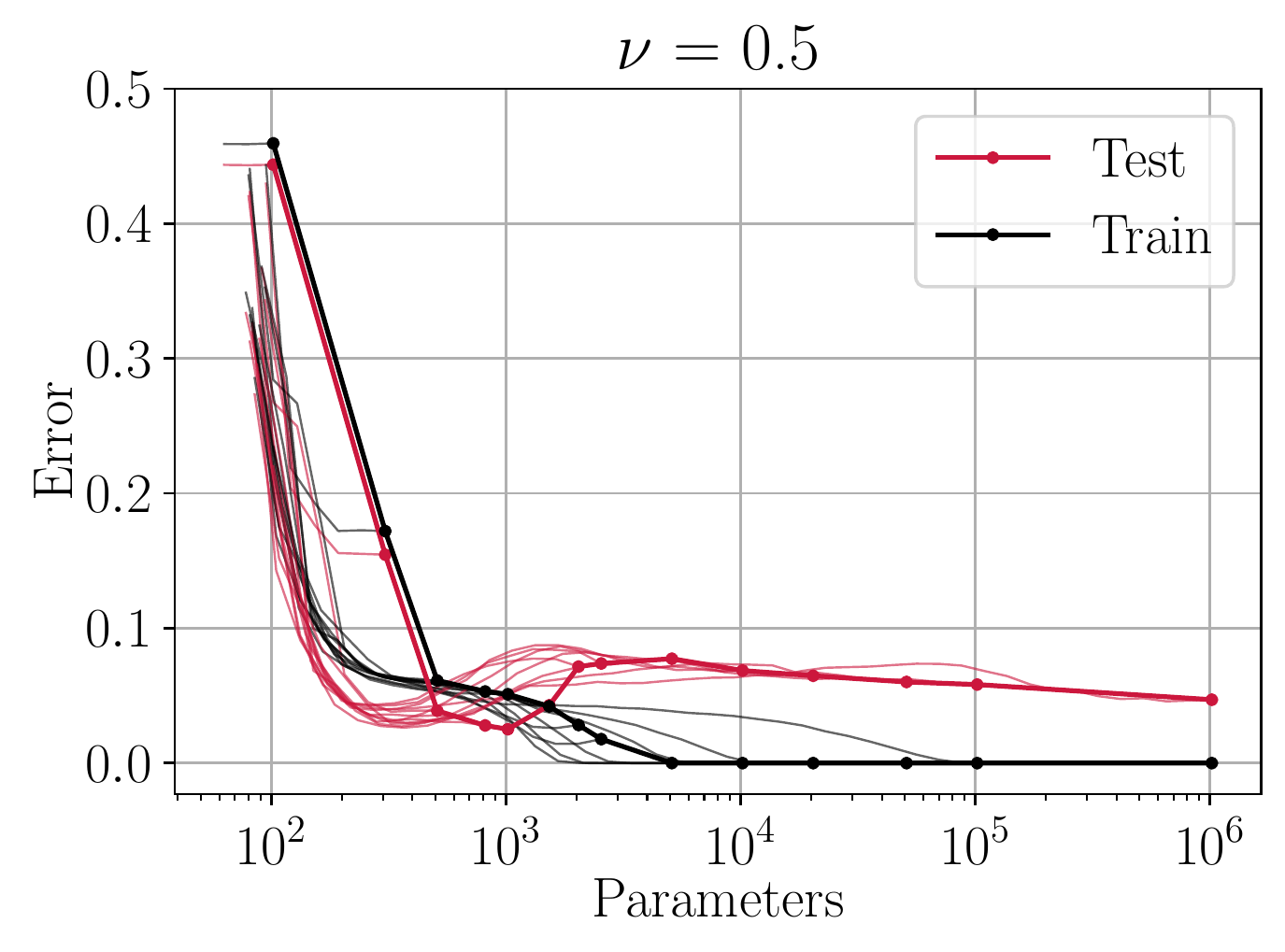}
    \includegraphics[width=.4\linewidth]{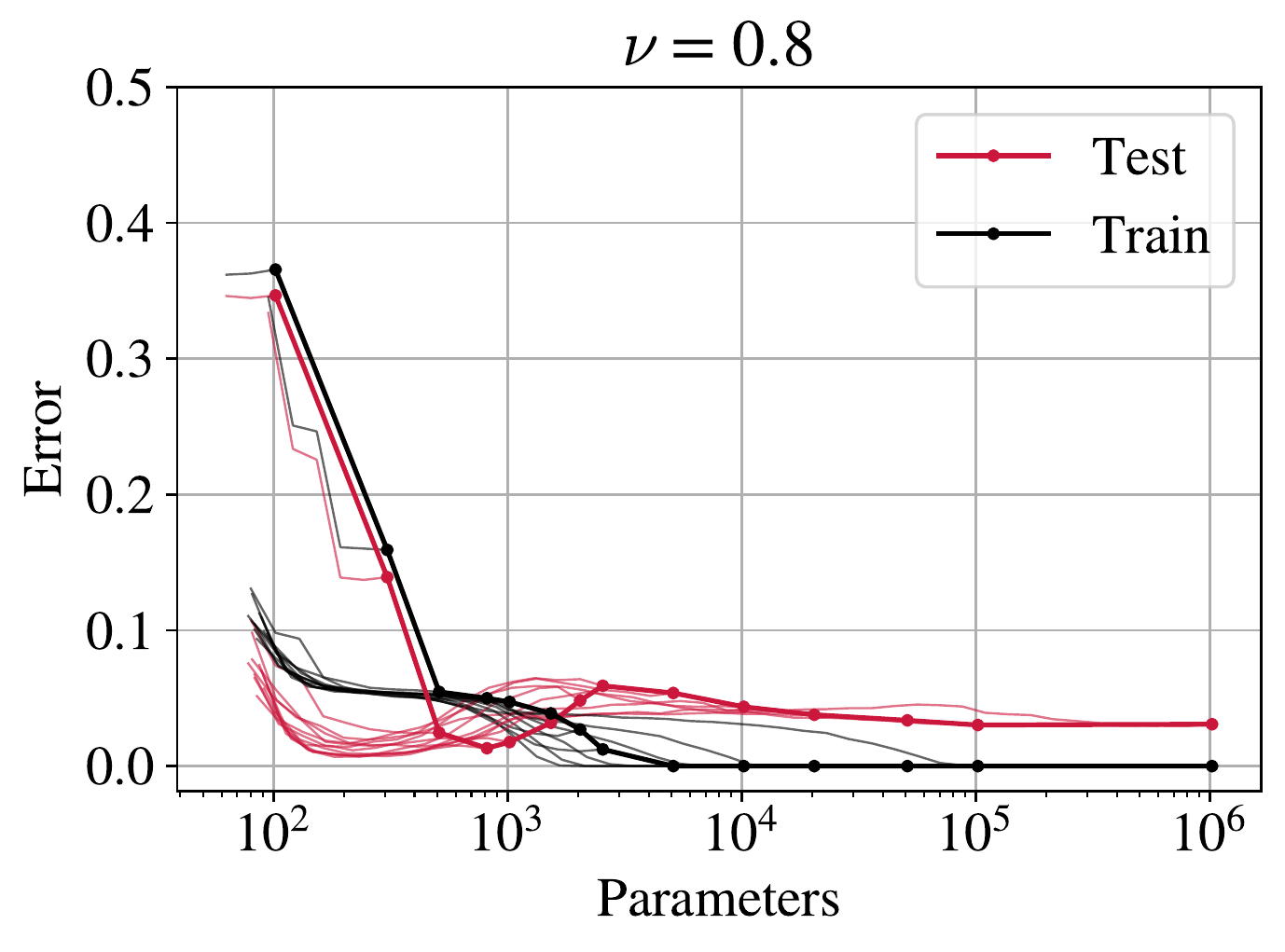}
    \caption{The first two principal components (PCs) of data generated using the linear datasets along with the double descent curves.}
    \label{fig:linear_ds}
\end{subfigure}
\begin{subfigure}{\textwidth}
\centering
    \begin{subfigure}{.19\textwidth}
    \centering
        \includegraphics[width=\linewidth]{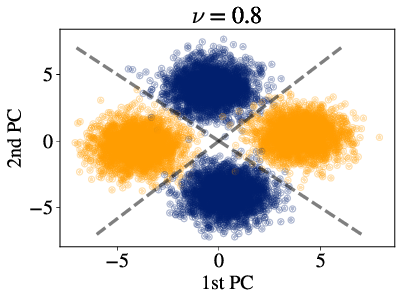}\\
        \includegraphics[width=\linewidth]{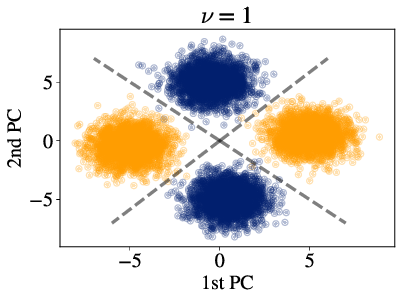}
    \end{subfigure}
    \includegraphics[width=.4\linewidth]{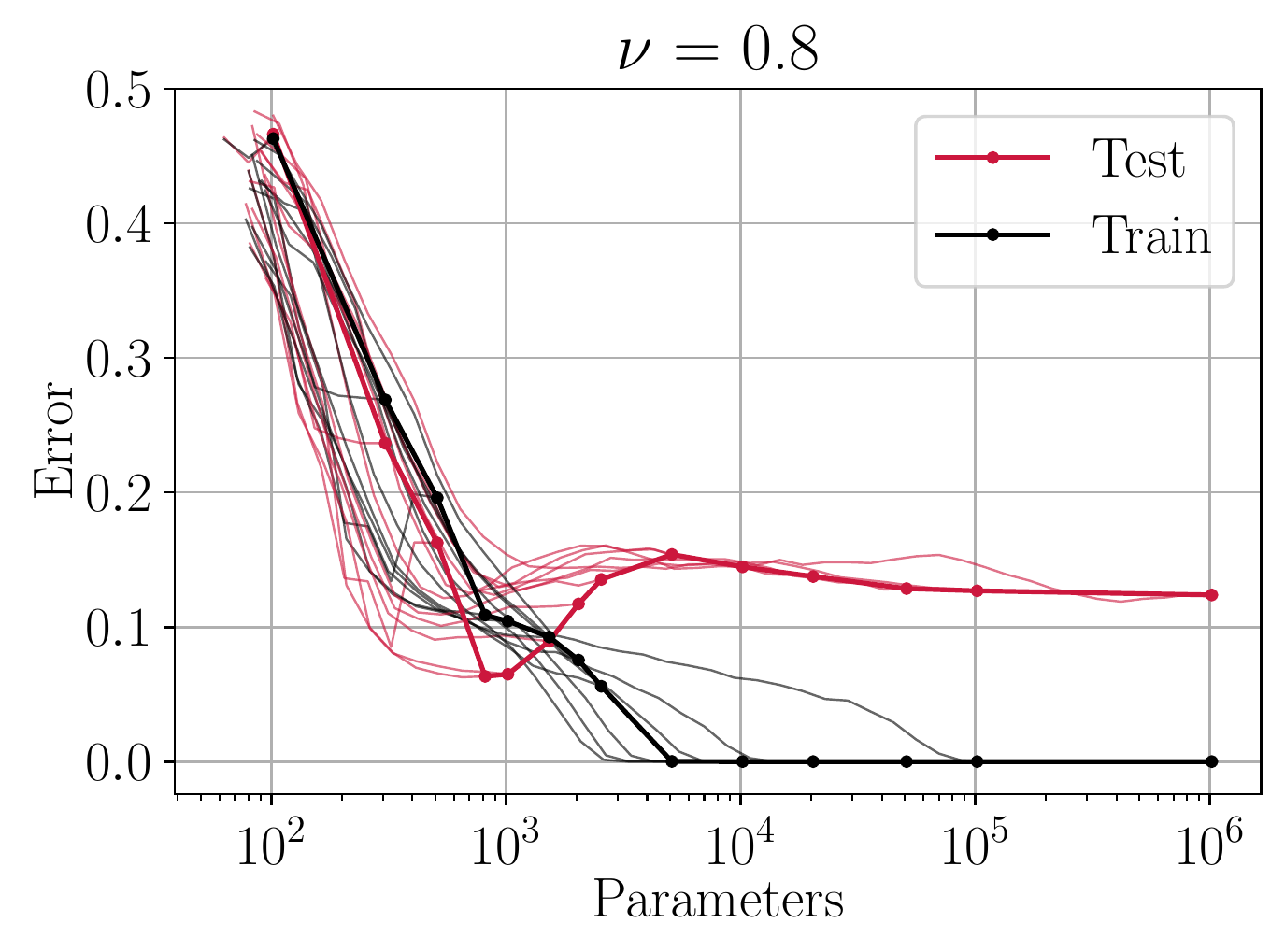}
    \includegraphics[width=.4\linewidth]{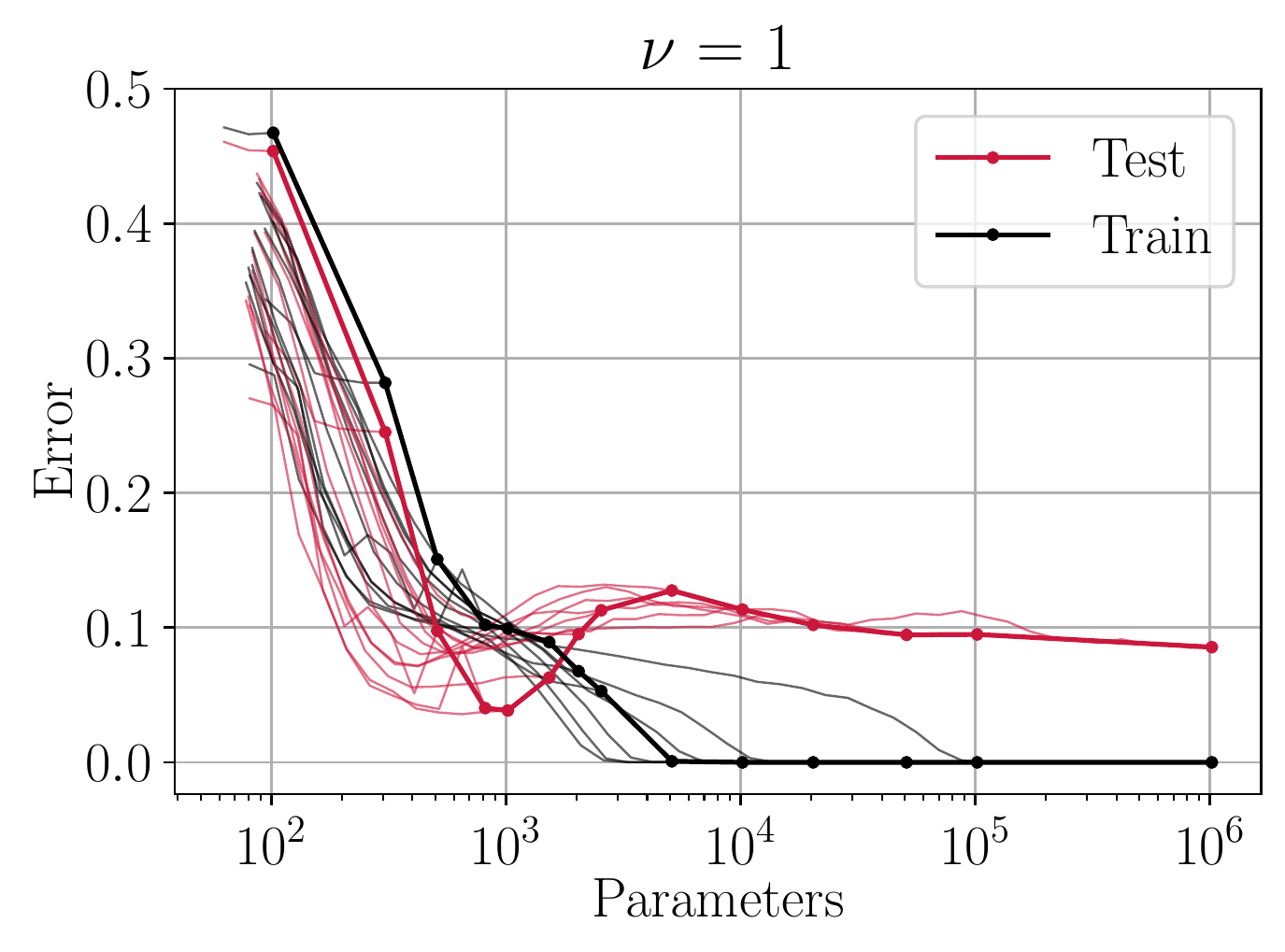}
    \caption{The first two principal components (PCs) of data generated using the XOR datasets along with the double descent curves.}
    \label{fig:XOR_ds}
\end{subfigure}
\caption{Average over 5 replicates of train and test error for models with different sizes, after numerous iterations of pruning, on the (a) linear, and (b) XOR datasets as the distance between clusters $\nu$ is varied. The red bold line with dots shows the traditional double descent curve, while the thinner lines represent the test error reached after pruning iterations of the initial models (sparse double descent). Similarly, black lines show train error both for the full and pruned models.}
\label{fig:all_ds}
\end{figure*}

\paragraph{Computational costs:} Our analysis requires us to work in the double descent paradigm, which allows us to properly define a best pruned model. Unfortunately, this implies significant computational costs as we have to perform the full sparse double descent analysis. For example, \cref{fig:sdd_n20,fig:all_ds,fig:mnist_app} required us to train approximately 300 different models and close to 1000 models when including replicates. Nevertheless, we believe that our extensive investigations using the mixture classification task and different architectures for MNIST and CIFAR-10 lay a solid foundation for the phenomenon described in the paper.

\paragraph{Terminology:} The results in \cref{fig:sdd_n20} show that repeated pruning of different-sized models produces sparse double descent curves that achieve the lowest test error within a small range of parameters. To facilitate further discussion on this phenomenon, we define \emph{best pruned model} and \emph{effective number of parameters}.

\begin{definition}[Best pruned model]
\label{def:best_pruned_model}
    Let $E(\cdot)$ denote the test error (or 0-1 loss) of a model. For a trained model $f_w$, test dataset $\mathbb{S}_\mathrm{test}$, and pruning method $\phi$, the best pruned model $f_w^\mathrm{bp}$ is defined as
    \[ f_w^\mathrm{bp} = \argmin_{w}\ E(\phi(f_w, \mathbb{S}_\mathrm{test})).\]
\end{definition}

These best pruned models have low generalisation error but unlike their overparameterized parent models, they have a non-zero error on the training data. 

\begin{definition}[Effective number of parameters]
\label{def:neff}
    The number of non-zero weights in the best pruned models $f_w^\mathrm{bp}$ obtained by pruning an overparameterised model. 
\end{definition}

By pruning overparameterised models of different sizes we can obtain best pruned models that have similar accuracy but a slightly different effective number of parameters.

\begin{table*}[!ht]
    \centering
    \caption{Number of parameters and test error for unpruned (full) and best pruned models for MNIST and CIFAR-10. Average values over 3 replicates are reported. We observe that a $\num{200}\times$ increase for the full models results in only a $\sim\num{3.5}\times$ increase in the number of parameters for the best pruned models. Notice also that the error achieved by pruned models appears insensitive to the error rate of the original full model, i.e. even models with poor generalisation can be rescued by pruning.}
    \label{tab:summary_params}
    \begin{tabular}{rr|rr||rr|rr}
       \multicolumn{4}{c||}{MNIST (3 layer FC)} & \multicolumn{4}{c}{CIFAR-10 (ResNet-18)} \\ \hline
       \multicolumn{2}{c|}{Parameters} & \multicolumn{2}{c||}{Test error} & \multicolumn{2}{c|}{Parameters} & \multicolumn{2}{c}{Test error} \\ \hline
       Full & Pruned & Full & Pruned & Full & Pruned & Full & Pruned \\ \hline
        \num{45200} & \num{6061} & \num{0.105} & \num{0.048} & \num{128271} & \num{17211} & \num{0.143} & \num{0.141} \\
        \num{89400} & \num{4908} & \num{0.176} & \num{0.059} & \num{238155} & \num{16359} & \num{0.159} & \num{0.143} \\
        \num{266200} & \num{5987} & \num{0.163} & \num{0.071} & \num{238155} & \num{16359} & \num{0.159} & \num{0.143} \\
        \num{443000} & \num{6377} & \num{0.144} & \num{0.064} & \num{1689080} & \num{19647} & \num{0.187} & \num{0.130} \\
        \num{885000} & \num{10196} & \num{0.123} & \num{0.084} & \num{3798420} & \num{35028} & \num{0.165} & \num{0.128} \\
        \num{4421000} & \num{10683} & \num{0.099} & \num{0.089} & \num{10546700} & \num{49799} & \num{0.146} & \num{0.128} \\
        \num{8841000} & \num{17094} & \num{0.091} & \num{0.097} & \num{23725050} & \num{57356} & \num{0.137} & \num{0.129} \\
    \end{tabular}
\end{table*}

\section{Results}%
\label{sec:res}
We first confirm that fully-connected neural networks exhibit both traditional double descent and sparse double descent on the mixture classification tasks: \cref{fig:all_ds} shows the same trend in the effective number of parameters as we observed in \cref{fig:sdd_n20}. As expected, the test error decreases with the distance between cluster means $\nu$ (see \cref{fig:neff_comparison}).

\begin{figure*}[!ht]
    \centering
    \begin{subfigure}{.45\textwidth}
        \includegraphics[width=\linewidth]{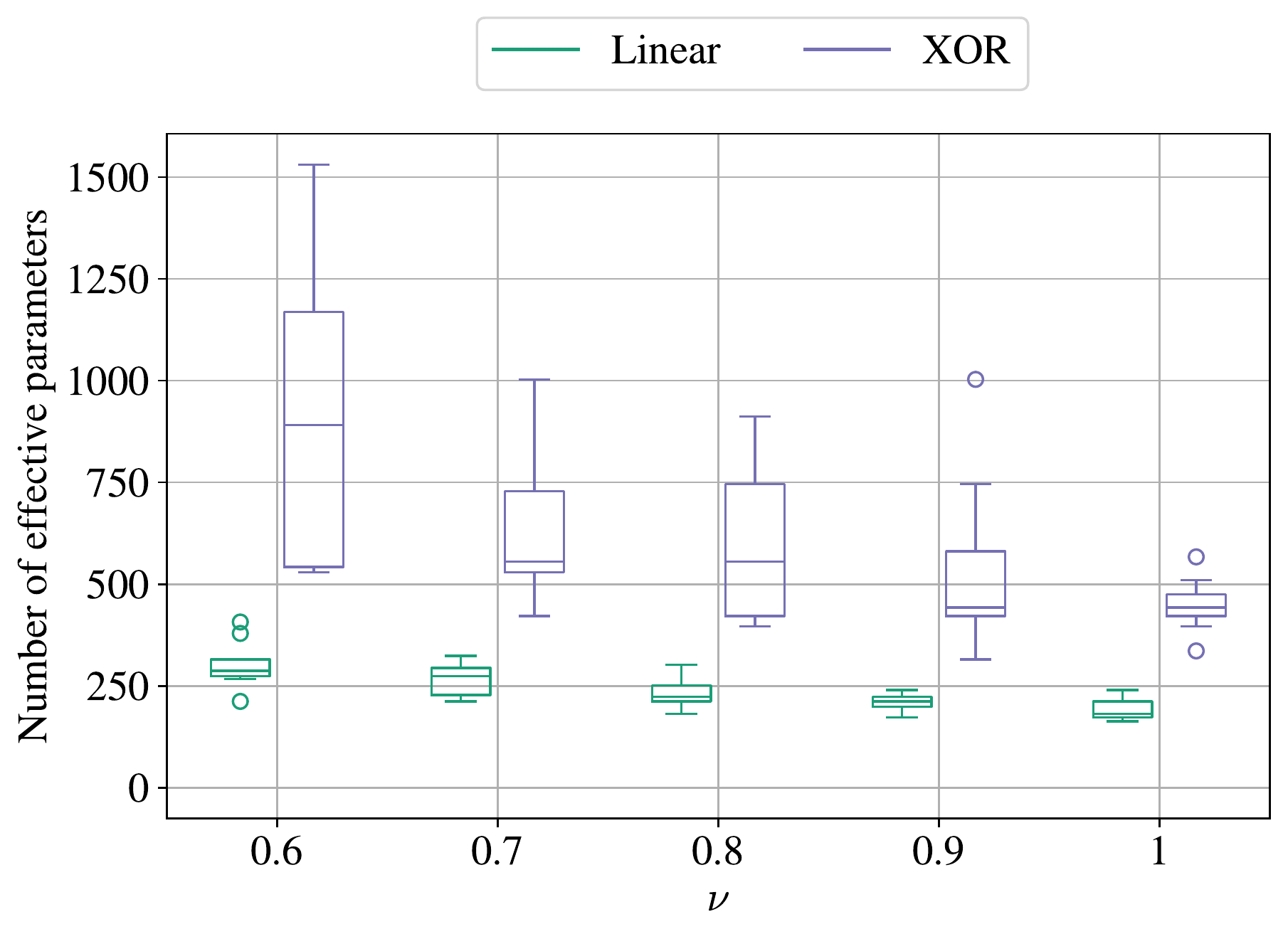}
        \caption{Comparing the effective number of parameters in the two mixture classification datasets for different values of $\nu$.}
        \label{fig:neff_comparison_datasets}
    \end{subfigure} \hspace{.5cm}
    \begin{subfigure}{.45\textwidth}
        \includegraphics[width=\linewidth]{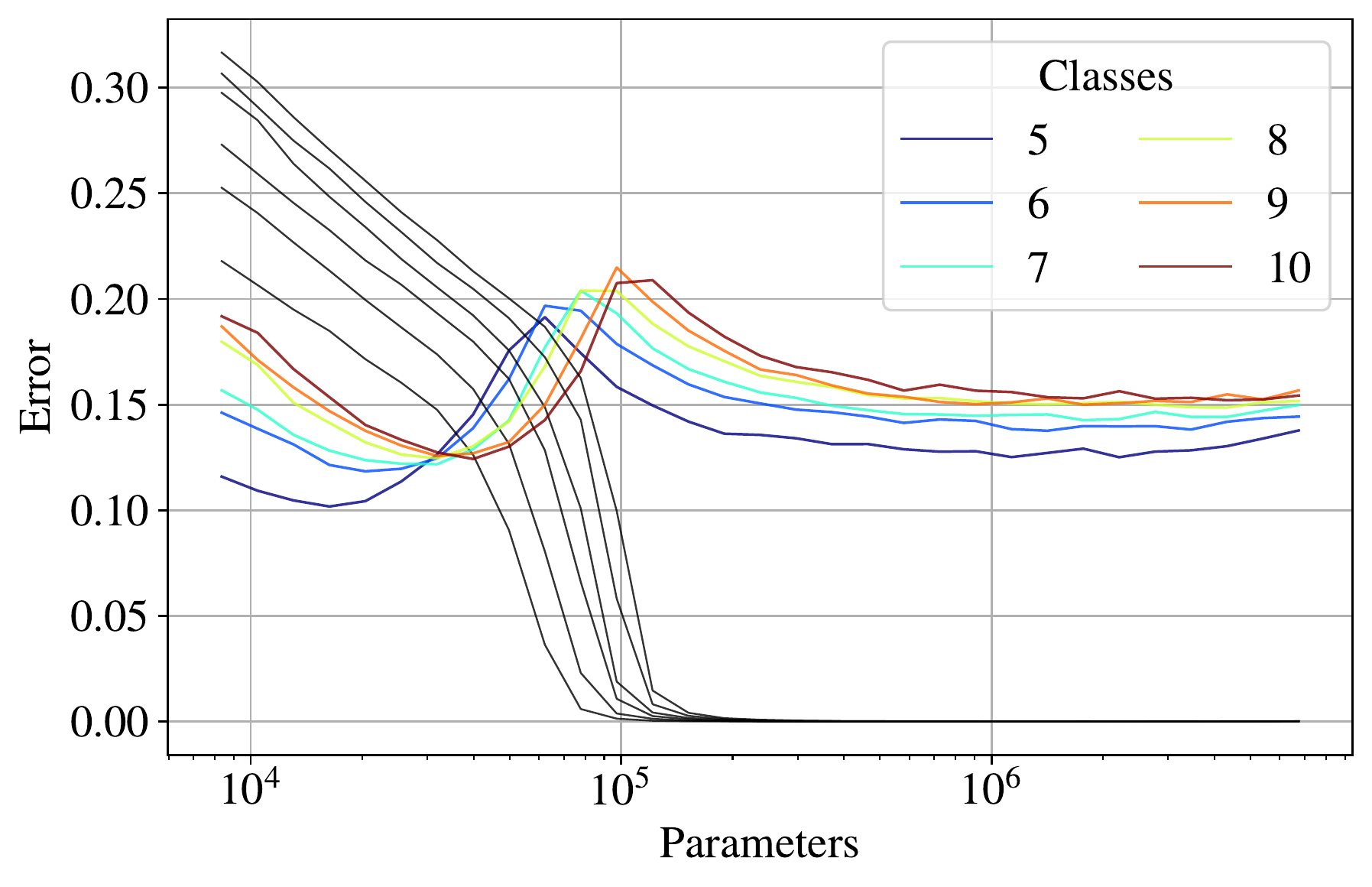}
        \caption{Sparse double descent curves on subsets of CIFAR-10 dataset show that effective number of parameters is related to task difficulty.}
        \label{fig:cifar_sub}
    \end{subfigure}
    \caption{Our analysis shows that the effective number of parameters correlates with task difficulty: (a) the effective number of parameters required for the XOR dataset is approximately twice that of the linear dataset, which reflects the higher complexity of the XOR task, and (b) decreasing the number of classes in the CIFAR-10 dataset allows IMP to find models with fewer effective number of parameters.}
\end{figure*}

Next, we considered the effective number of parameters in the best pruned models (cf.~\cref{def:best_pruned_model}). Focusing on the linear dataset in \cref{fig:linear_ds}, we see that the pruned models (thin red lines) generally achieve the lowest test error with roughly \num{500} parameters. This number of parameters is optimal for starting networks of different size, which we use to determine the effective number of parameters for that dataset (see \cref{def:best_pruned_model,def:neff}). For networks trained on the linear dataset, we find that the effective number of parameters, i.e.~the number of parameters in the best pruned models, ranges from $\sim \num{200}$ to $\sim \num{500}$ for different starting models and for different values of $\nu$ (see \cref{fig:linear_neff} for the full distribution). We find the same behaviour for the XOR dataset, \cref{fig:XOR_ds}, but with a higher number of parameters (between $\num{250}$ and $\num{1000}$) than for the linear dataset with the same $\nu$.

The double descent and sparse double descent curves for MNIST and CIFAR-10 datasets can be seen in \cref{fig:sdd_n20,fig:mnist_app}. \Cref{tab:summary_params,tab:summary_mnist} show the number of parameters and test errors for the full/unpruned models and the corresponding best pruned models. Notice, importantly, that the best pruned models do not interpolate, so they exhibit a significant improvement in generalisation gap. Focusing on the number of parameters, we observe that a $\num{200}\times$ increase for the full models results in only a $\sim\num{3.5}\times$ increase in the number of parameters for the best pruned models. Comparing the test errors, a surprising finding is that sparse subnetworks with low test error exist inside the unpruned models that lie at the interpolation regime of the double descent curve. This means that even though the unpruned models have high test errors, they can be pruned using IMP to achieve much lower test errors. Pruning thus acts as a type of regularisation in this case, which is known to mitigate the peak of the test accuracy at the interpolation threshold \citep{belkin2019reconciling}. 

Our findings for CIFAR-10 similarly show that the best pruned models have better generalisation than the unpruned counterparts (see \cref{tab:summary_params}). Similar to MNIST, we observe that the number of parameters in the best pruned models increases with the size of the original model but at a much slower rate. Comparing the effective number of parameters for MNIST and CIFAR-10, one can easily conclude that more parameters are required for CIFAR-10.

\begin{figure*}[!ht]
    \centering
    \begin{subfigure}{.3\textwidth}
        \includegraphics[width=\linewidth]{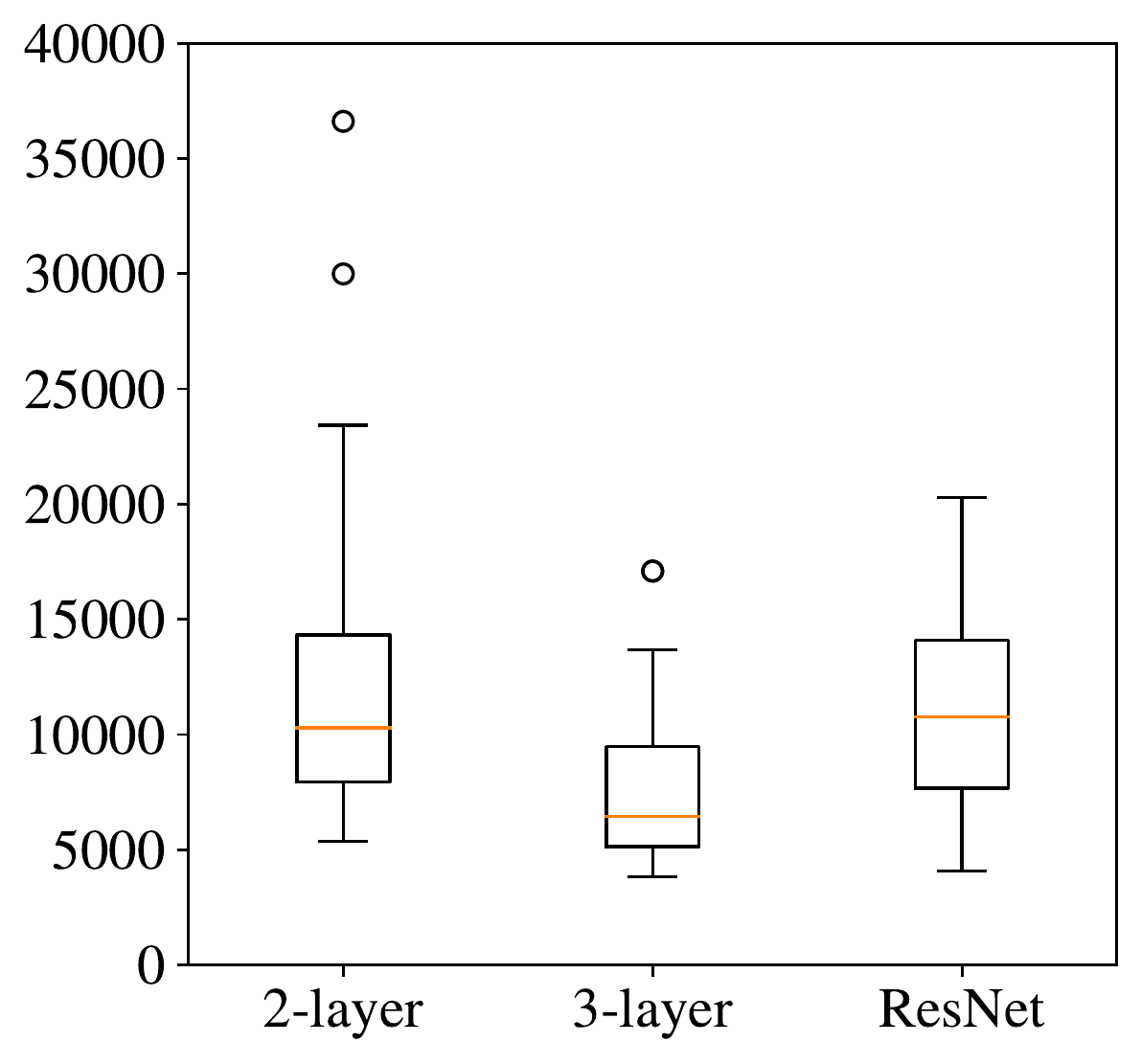}
        \caption{Effective number of parameters for different models trained on the MNIST dataset.}
        \label{fig:mnist_params}
    \end{subfigure} \hspace{.5cm}
    \begin{subfigure}{.45\textwidth}
        \includegraphics[width=.9\linewidth]{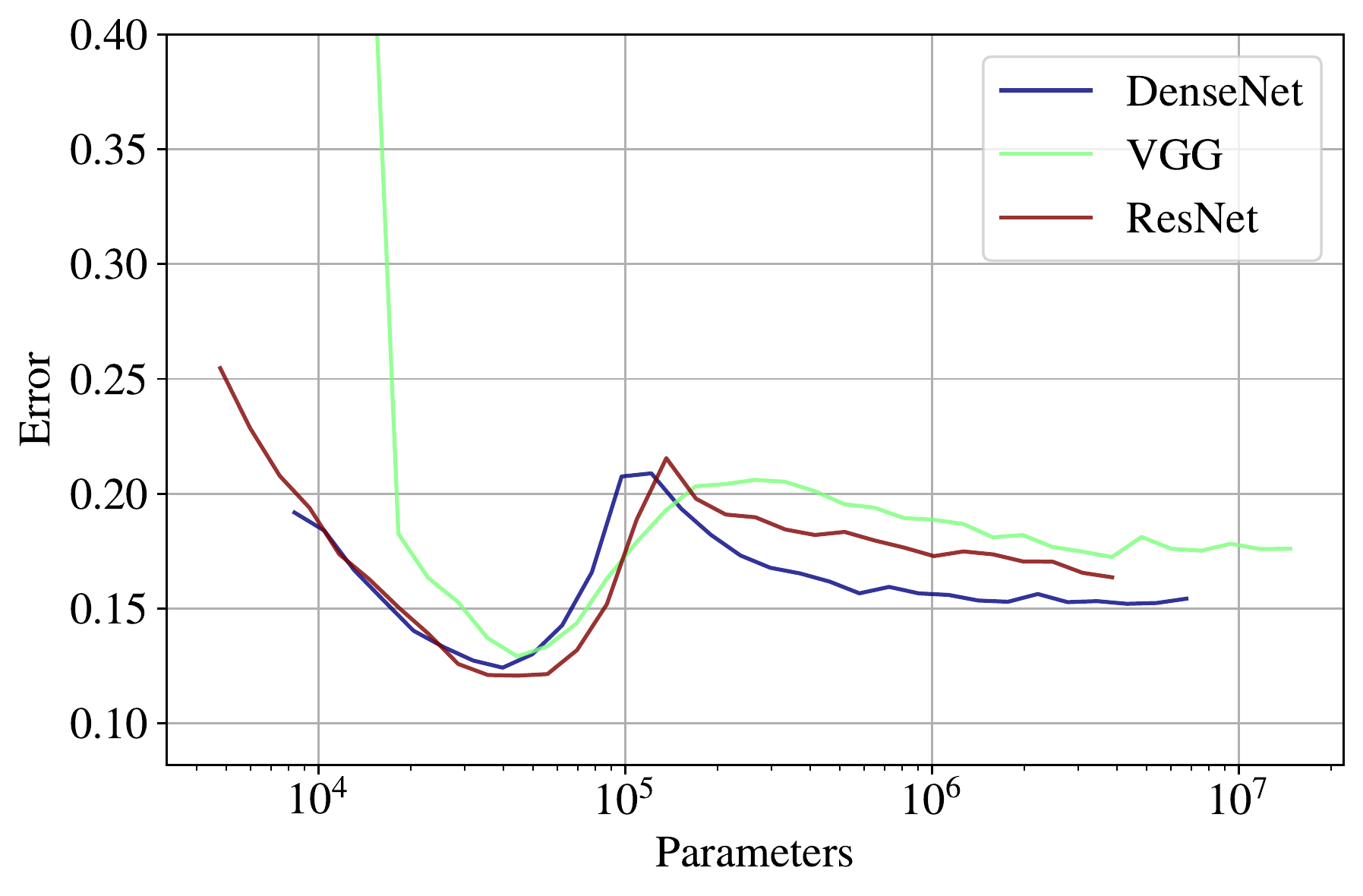}
        \caption{Sparse double descent curves obtained on pruning different convolutional neural networks for CIFAR-10 dataset.}
        \label{fig:cifar_mods}
    \end{subfigure}
    \caption{Comparing the effective number of parameters across different neural network architecture for MNIST and CIFAR-10 datasets.}
\end{figure*}

\paragraph{The effective number of parameters correlates with task difficulty:} 
As illustrated in the data plots of \cref{fig:all_ds}, optimally separating the linear dataset requires a plane, whereas two planes are needed for separating the classes in the XOR dataset. Thus, one would expect that the effective number of parameters would double going from the linear to XOR datasets. Interestingly, this is what we observe in \cref{fig:neff_comparison_datasets}, even though the best pruned models are obtained from full models of the same size. The pruning approach thus suggests a way to quantify the effective number of parameters in deep, over-parameterised neural networks. Can we extend this approach to measuring task difficulty to more realistic datasets and convolutional networks? The results in \cref{fig:cifar_sub} and \cref{tab:cifar_sub} show that for a fixed ResNet model, as we decrease the number of classes in the CIFAR-10 dataset (or equivalently make the task easier for an overparameterised model), the effective number of parameters reduces for a smaller classification task. This further affirms our observation on the mixture classification task that the effective number of parameters is related to the difficulty of the task. Interestingly, the effective number of parameters does not decrease linearly with the number of classes.

\paragraph{The effective number of parameters is comparable across architectures:}
How does the choice of neural network architecture impact the effective number of parameters for a given dataset? We compare the size of best pruned models obtained for the MNIST dataset using three different neural network architectures: two-layer FCN, three-layer FCN, and ResNet-6, in \cref{fig:mnist_params}. We find that although the test error of the best pruned models for the different architectures is different (see \cref{tab:summary_params,tab:summary_mnist}), the effective number of parameters are fairly comparable. This provides additional empirical evidence that our procedure might be invariant to the neural network architecture and reflects a notion of the `difficulty' of the classification task, something that we already noted in the mixture classification task. This observation also carries forward to different convolutional neural network architectures on CIFAR-10. \Cref{fig:cifar_mods} shows the sparse double descent curves obtained for DenseNet, VGG, and ResNet. The effective number of parameters is similar across the three different architectures: \num{44590} for DenseNet, \num{44468} for VGG, and \num{39483} for ResNet.

\begin{figure*}[!ht]
    \centering
    \begin{subfigure}{.44\linewidth}
        \includegraphics[width=\linewidth]{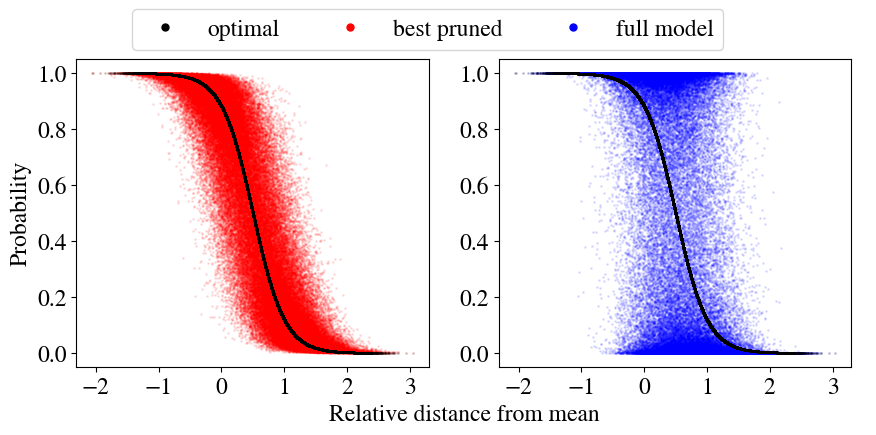}    
        \caption{Comparing the decision boundaries learned by the best pruned and full models we find that best pruned models are more aligned with the optimal classifier.}
        \label{fig:dist_prob}
    \end{subfigure}\hfill
    \begin{subfigure}{.55\linewidth}
        \includegraphics[width=.49\textwidth]{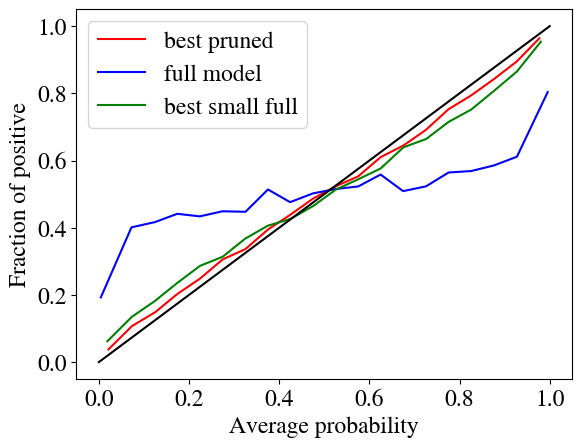} \hfill
        \includegraphics[width=.49\textwidth]{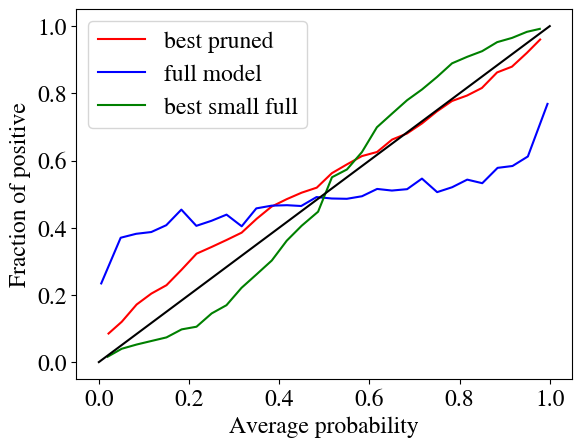} 
        \caption{Calibration curves on the linear (left) and XOR (right) datasets show that best pruned models and small full models are better calibrated, whereas the full models are overconfident and poorly calibrated.}
        \label{fig:calib_mixt}
    \end{subfigure}\\
    \begin{subfigure}{\textwidth}
        \includegraphics[width=.49\textwidth]{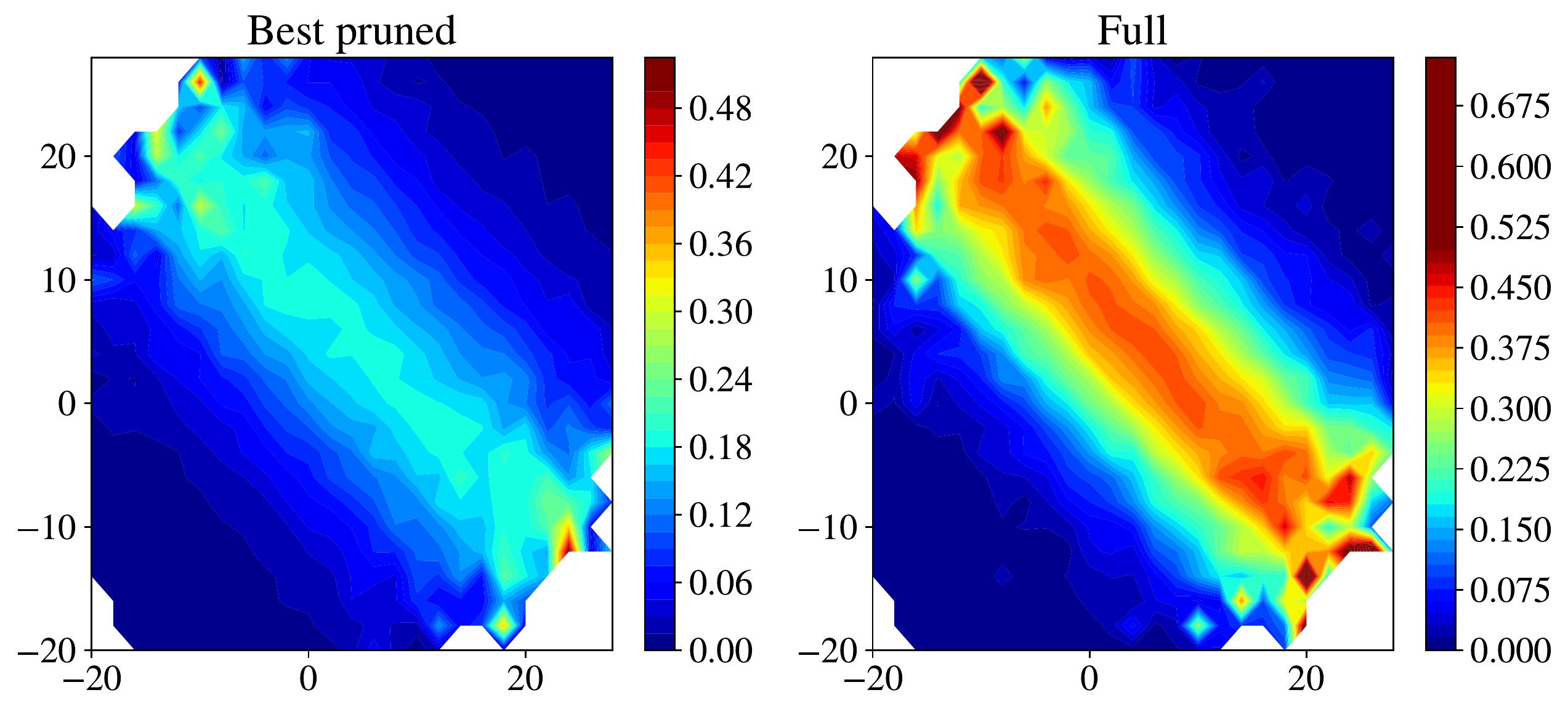} \hfill
        \includegraphics[width=.49\textwidth]{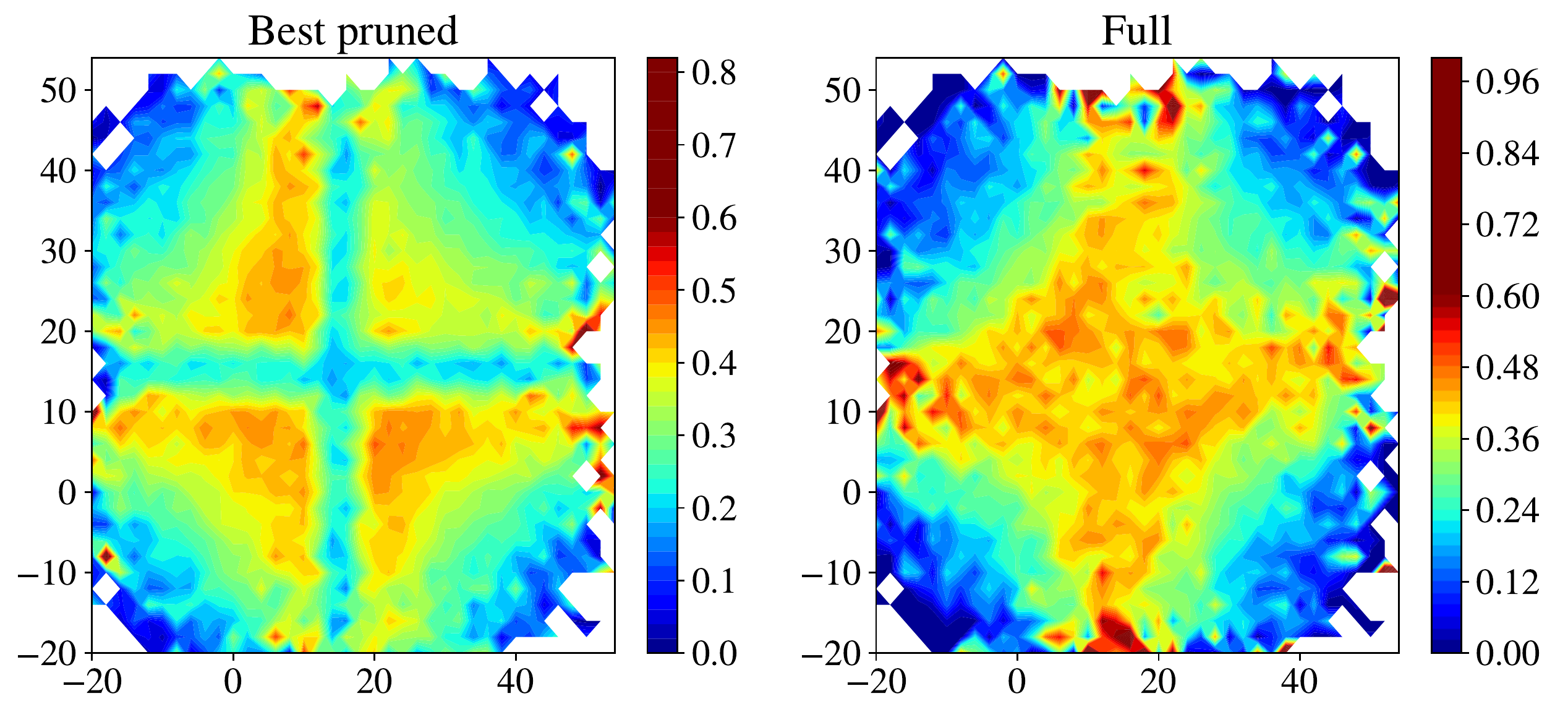}
         \caption{Visualising the absolute difference between the predicted class probabilities and the optimal probabilities for the linear (left) and XOR (right) datasets projected on a 2D plane. Lower values (darker blue colours) are preferred.}
        \label{fig:db_pdiff}
    \end{subfigure}
    \caption{Analysis of prediction uncertainty in the mixture classification task. We find that the best pruned models are better at capturing uncertainty, whereas the overparameterised models tend to be overconfident in their predictions.}
    \label{fig:mixt_analysis}
\end{figure*}

\subsection{Pruned models better capture uncertainty}
\label{sec:model_analysis}
We first use the mixture classification task to understand the architectural bias induced by pruning. For the results presented in \cref{fig:mixt_analysis}, we evaluate the models with $\num{500}$ neurons in the hidden layer for the linear ($\nu=0.2$) and XOR ($\nu=0.6$) datasets. Similar results are obtained for different starting models and other values of $\nu$. To understand what differentiates the pruned models we analyse the decision boundary learned by these models. We sample $\num{100000}$ points from the original data distribution and compute the corresponding probabilities for the models of interest along with the conditional class probabilities. In \cref{fig:dist_prob}, we plot the probability of data belonging to class $y=0$ as a function of the relative distance between the centres of the two clusters (located at $x=0$ and $x=1$) for the linear dataset. As expected, the optimal probabilities are given by a logistic function shown using the black dots. Comparing the best pruned and full models, we can conclude that the function learnt by the best pruned model is much closer to the true class conditional probability used to generate the data, whereas the full overparameterised model is overconfident in predicting either class. 

In \cref{fig:db_pdiff}, we visualise the absolute difference between the predicted probabilities and the conditional class probabilities relative to the position of each point projected on a 2D plane. A value close to $0$ on the colour scale in \cref{fig:db_pdiff} implies that the model has correctly estimated the conditional probability. One would expect that a model would confidently predict the correct class farther from the boundary and make errors near the boundary. This intuition is clearly illustrated using the results for the linear dataset. Strikingly, the full model makes confident predictions near the boundary, thus resulting in a value of $\approx 0.45$ shown using the red colour in \cref{fig:db_pdiff}. In the XOR dataset, the best pruned model is more uncertain in its predictions, especially near the boundary, but is better than the full model. This is further illustrated using the calibration curves in \cref{fig:calib_mixt} where we see a near-perfect calibration in the best pruned models for both datasets, whereas the full models are overconfident and poorly calibrated. We also observe that the calibration of the best small full models is comparable to the best pruned models.

\begin{figure*}[!ht]
    \centering
    \begin{subfigure}{.39\linewidth}
        \includegraphics[width=\textwidth]{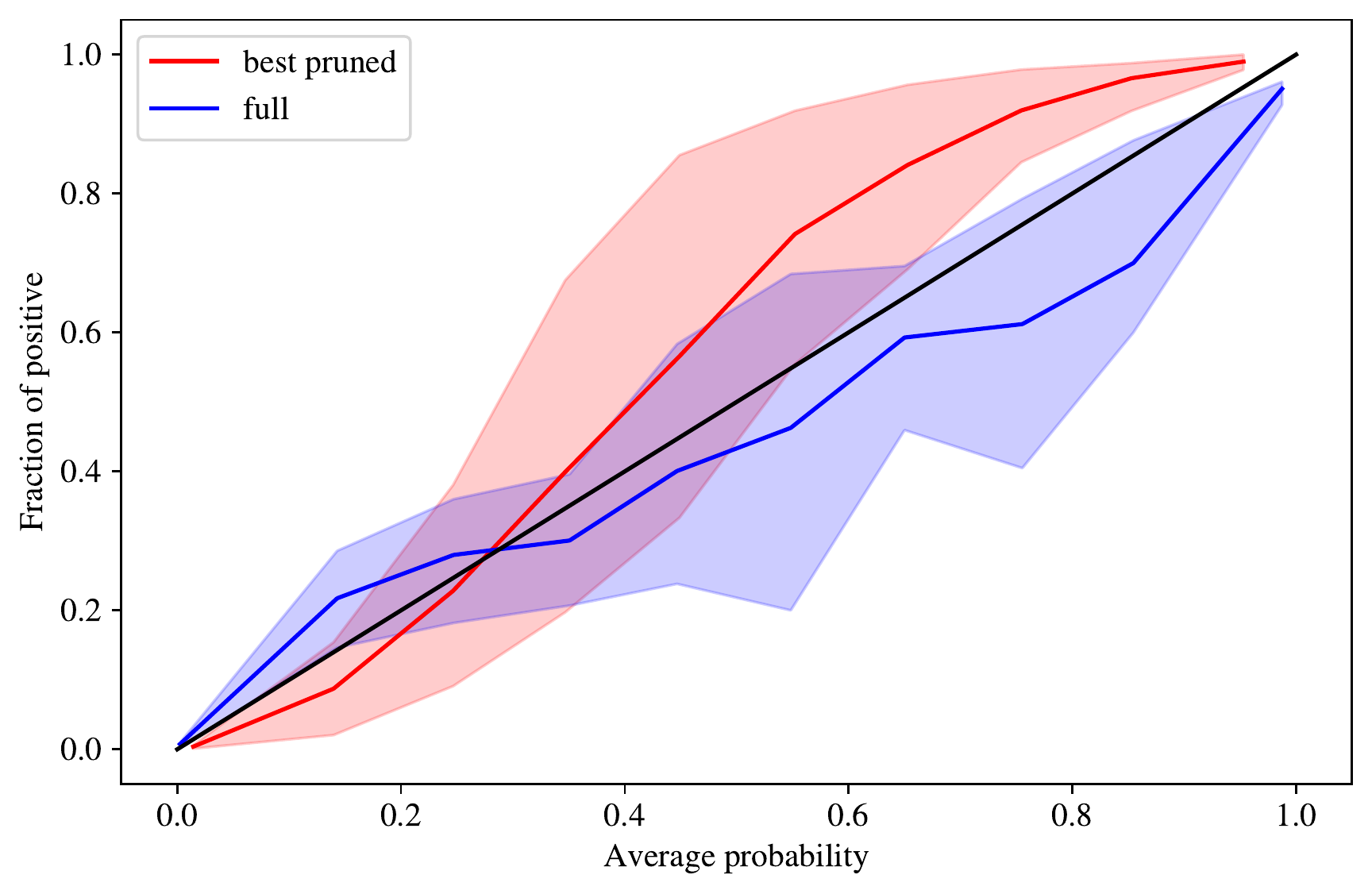}
        \caption{MNIST with two-layer FCN. ECE for pruned and full models are $0.114$ and $0.052$, respectively.} \label{fig:calib_mnist}
    \end{subfigure}\hfill
    \begin{subfigure}{.39\linewidth}
        \includegraphics[width=\textwidth]{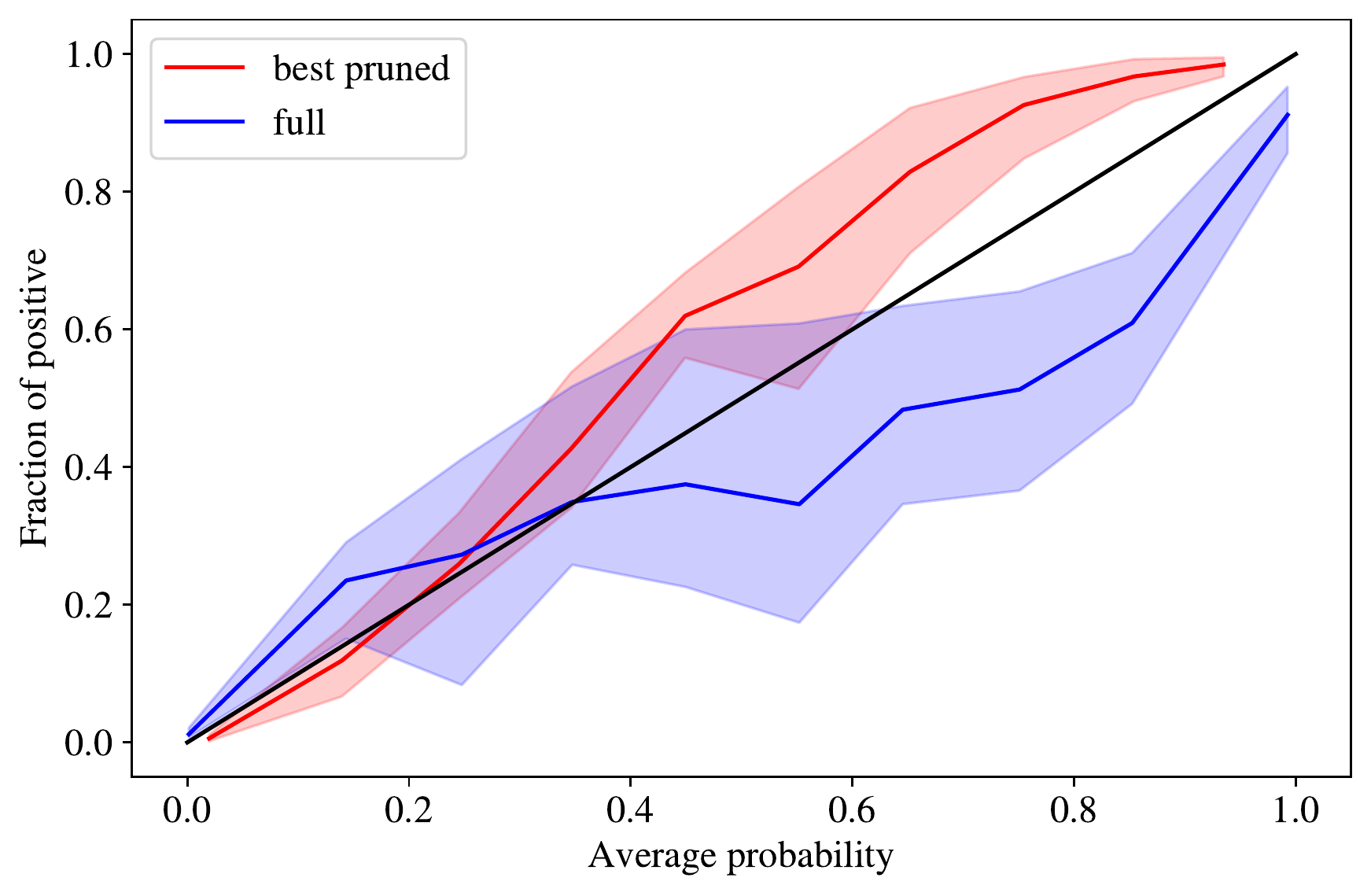}
        \caption{CIFAR-10 with ResNet-18. ECE for pruned and full models are $0.122$ and $0.101$, respectively.} \label{fig:calib_cifar}
    \end{subfigure}
    \begin{subfigure}{.2\linewidth}
        \centering
        \includegraphics[width=\linewidth]{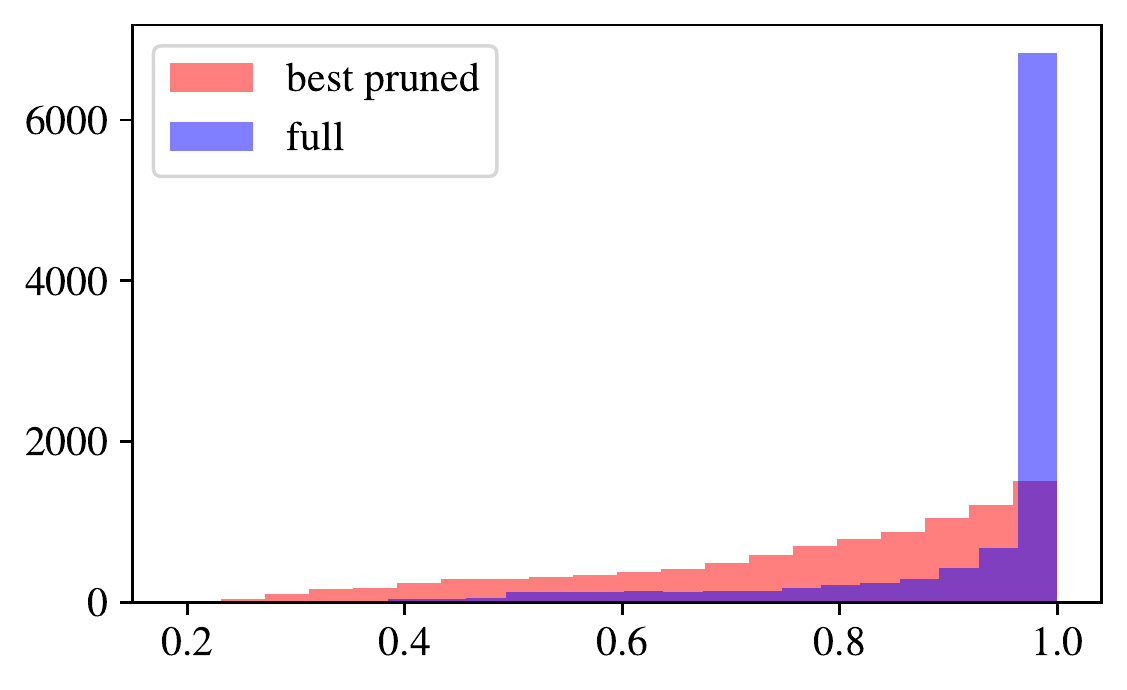}\\
        \includegraphics[width=\linewidth]{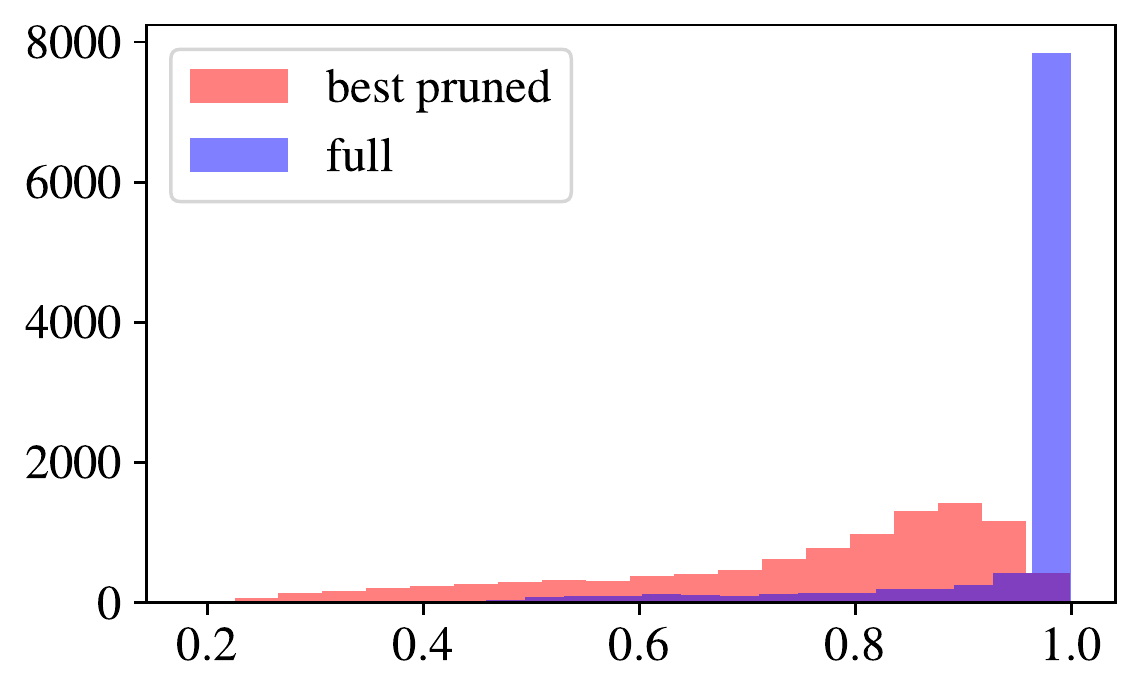}
        \caption{Probabilities of the predicted class.} \label{fig:p_hists}
    \end{subfigure}
    \caption{Class-averaged calibration curves for the best pruned and full models on (a) MNIST and (b) CIFAR-10 datasets show that the pruned models are underconfident while the full models are overconfident. The highlighted areas signify deviation between classes. (c) Probabilities of the predicted class for the MNIST (top) and CIFAR-10 (bottom) datasets.}
    \label{fig:calib_real}
\end{figure*}

\begin{figure*}[!ht]
    \centering
    \includegraphics[width=.49\linewidth]{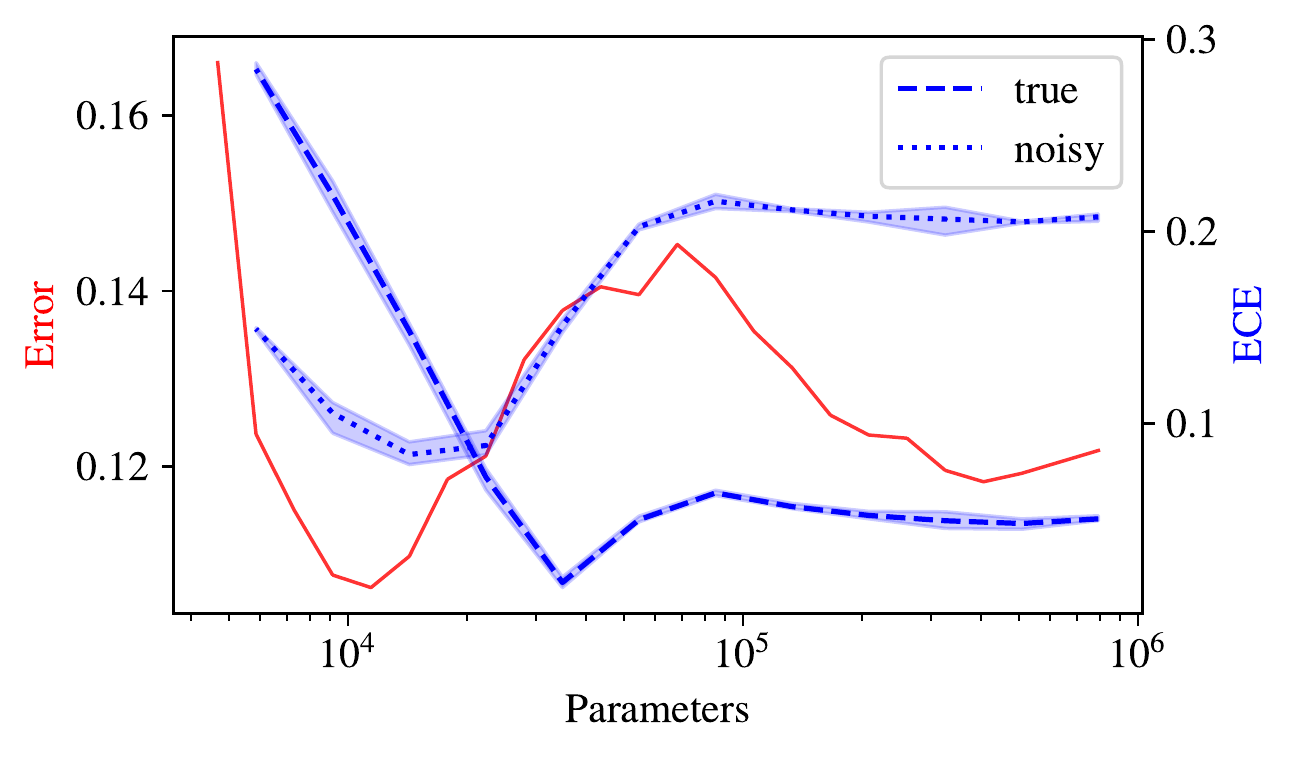}
    \includegraphics[width=.49\linewidth]{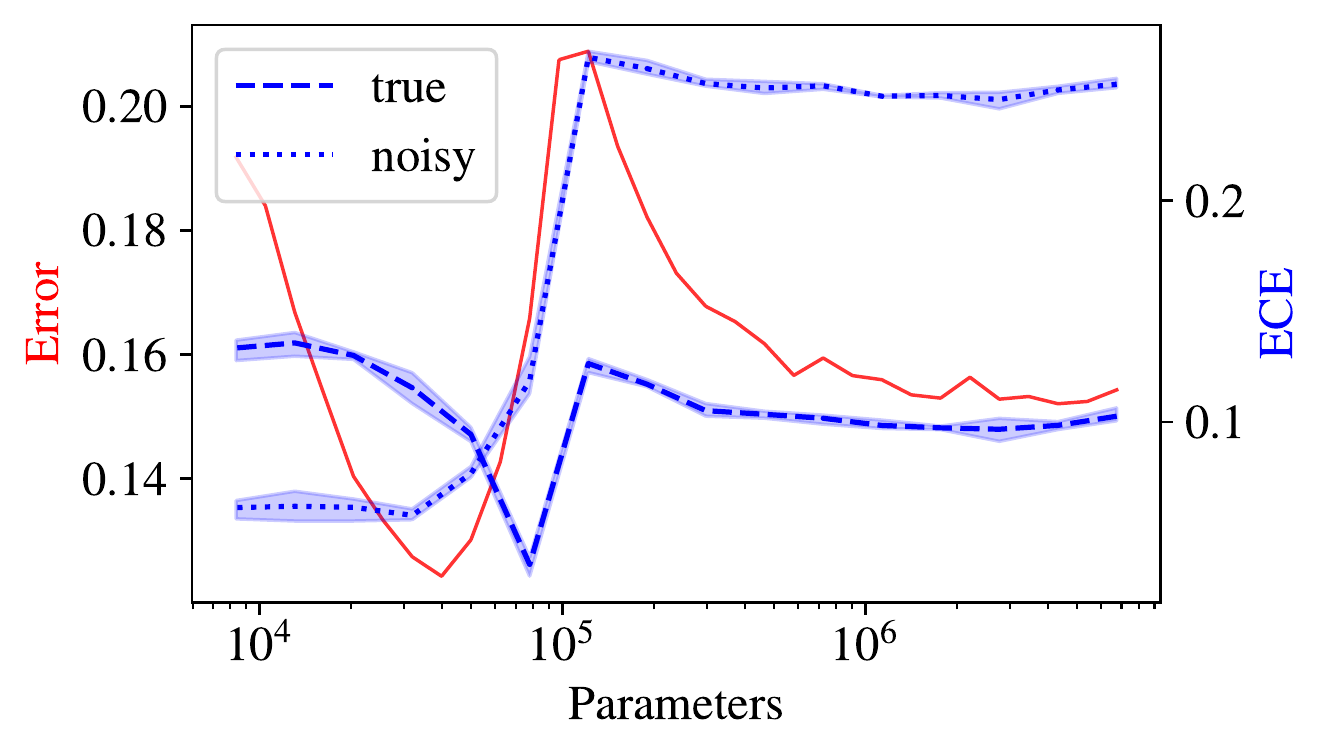}
    \caption{Sparse double descent (red) and expected calibration error (blue) on pruning two-layer FCN on MNIST (left) and ResNet-18 for CIFAR-10 (right). Highlighted area shows the deviation across three replicates. Comparing the calibration error for true and noisy labels we find that the best pruned models are optimally calibrated to noisy data.} \label{fig:ece}
\end{figure*}

The calibration curves for MNIST and CIFAR-10 in \cref{fig:calib_real} show that the curves for the best pruned models are usually above the black line, which means that they make the correct prediction while being uncertain. \Cref{fig:p_hists} illustrates this using the probabilities corresponding to the predicted class. This implies that even though the pruned models are accurate, they tend to be underconfident in their predictions while the full models are overconfident. This is at odds with our observation in the mixture classification task. On further analysis, we found that the best pruned models are calibrated to test data with label noise (see \cref{fig:calib_real_noise}). This is a reasonable outcome since the model was trained with label noise, which could introduce significant bias in the classification learned by the model.  

To understand how iterative pruning impacts calibration, in \cref{fig:ece} we plot the test error and expected calibration error~\citep{guo2017calibration} (ECE) for a fixed model. For both noisy and true labels, we observe that the ECE initially increases slightly on iterative pruning but then drops significantly. The major difference between the two scenarios is the location of the minimum. For true labels, the lowest ECE is observed for models that have high test error, whereas the best pruned models are better calibrated to data with noisy labels. This strange behaviour requires further investigation to understand the impacts of pruning on model calibration. Perhaps, a trade-off between test error and calibration error can be used to choose models that have low error and are well-calibrated.

\section{Conclusions}
\label{sec:conc}

The existence and identifiability of small pruned networks, which can still generalise as well as large, over-parameterised models, remains an empirically well-supported fact that still lacks a satisfactory theoretical explanation. In this paper, we provided a number of empirical observations which unveiled two intriguing characteristics of small pruned networks: (1) their number of parameters correlate with task difficulty and provide a measure of the effective number of parameters in large models, and (2) pruned models are less prone to overconfident predictions.

Working in the IMP framework, we consistently reproduce the sparse double descent phenomenon first observed by \citet{he2022sparse} in a number of synthetic and real datasets. While \citet{he2022sparse} focused on the sparse double descent of the large model, we observe that irrespective of the size (and test accuracy) of the model from which we start the sparse double descent, the resulting optimal pruned models all achieve a similar generalisation error and have comparable sizes. The size of the best pruned models (defined by the number of retained parameters) appears to correlate well with natural notions of task complexity, both in real and simulated datasets.

We then analysed more in-depth the functions learned by these small networks: on simulated data, where we know the actual class conditional probability functions, we observe that pruned models capture much better the true underlying function, whereas full models tend to be overconfident. On real data, we again verify that full models are more confident than pruned models, which in this case tend to be slightly under-confident. A possible explanation is that, as customary in double descent studies, our models are trained on data with label noise, which impacts their calibration on data with true labels. Indeed, we find that pruned models are nearly optimally calibrated to data with label noise, while full models are once again over-confident.

Our results are somewhat at odds with a recent claim by \citet{yang2023theoretical}, which showed evidence of over-confidence of pruned models in one example (ResNet-50 on CIFAR-100). A possible reason for this discrepancy is that the pruning algorithm employed in \citep{yang2023theoretical} is greedy and does not require re-training from initialisation. Additionally, their setup did not involve training on noisy labels; both of these observations might explain the different conclusions reached by that study. Indeed, these considerations point to a non-trivial interaction between pruning algorithm, training procedure and statistical characteristics of the resulting pruned models. Exploring such questions, both theoretically and empirically, might finally shed light on the unreasonable success of pruning large neural networks.

In the future, it would be interesting to see how other pruning procedures impact our findings and extend our study to larger datasets like ImageNet. It would be interesting to see if these findings apply to other architectures like auto-encoders and recurrent neural networks. Finally, our findings highlight the importance of the sparse double descent thus encouraging theoretical work to understand the phenomenon.

\begin{contributions} 
Viplove Arora, Sebastian Goldt, and Guido Sanguinetti conceived the idea and wrote the paper. Daniele Irto performed the experiments for the mixture classification task. Viplove Arora performed the rest of the experimental analysis.

\end{contributions}

\begin{acknowledgements} 
We acknowledge co-funding from Next Generation EU, in the context of the National Recovery and Resilience Plan, Investment PE1 – Project FAIR ``Future Artificial Intelligence Research''. This resource was co-financed by the Next Generation EU [DM 1555 del 11.10.22]. The views and opinions expressed are only those of the authors and do not necessarily reflect those of the European Union or the European Commission. Neither the European Union nor the European Commission can be held responsible for them. Viplove Arora acknowledges partial support from Assicurazioni Generali through a liberal grant.
\end{acknowledgements}

\bibliography{arora_437}

\setcounter{table}{0}
\renewcommand{\thetable}{S\arabic{table}}%
\setcounter{figure}{0}
\renewcommand{\thefigure}{S\arabic{figure}}%
\setcounter{equation}{0}
\renewcommand{\theequation}{S\arabic{equation}}
\setcounter{section}{0}
\renewcommand{\thesection}{S\arabic{section}}
  
\onecolumn
\appendix
\section{Mixture Classification Datasets}
\label{app:mixt_class}
A common technique in the study of neural network theory is to operate in a controlled setting, where models and data are simplified to allow analytical computations for otherwise intractable objects. An example of such an approach is the teacher-student framework \citep{seung1992statistical}, where the dataset is created by a neural network built for that purpose (teacher) and another neural network is tasked with learning on that data (student). For our analysis, we create two types of balanced binary classification datasets using Gaussian mixtures with different characteristics and varying levels of difficulty. Using the mixture model allowed us to create datasets that are easy to understand and can be easily visualised in a two-dimensional space. 

We consider two settings for mixture classification. The \emph{linear} dataset consists of two clusters that can be separated using a linear function. The two clusters have different means $\mu_1 \neq \mu_2$ but same covariance $\Sigma_1 = \Sigma_2$. The \emph{XOR} dataset was created such that the resulting clusters are placed like the graphic representation of the XOR logical function. We assume that the difficulty of the classification task would increase going from linear to XOR dataset. Data was sampled using a Gaussian mixture model conditioned on a predefined label $y$. We also define a coefficient of separation $\nu$ to modulate the distance between clusters. 

To properly define the formulation of our mixture models, we use an approach similar to that used by \citet{refinetti2021classifying}. The data distribution for a single input sample $\mathbf{x} \in \mathbb{R}^D$ given class $y$ sampled uniformly at random is:

\begin{equation}
    p(\mathbf{x},y) = p(y) \; p(\mathbf{x}|y), \quad p(\mathbf{x}|y) = \sum_{\alpha \in \mathcal{S}^T(y)} \mathcal{P}_\alpha \, \mathcal{N}(\mathbf{\mu}_\alpha , \mathbf{\Sigma}_\alpha).
\end{equation}

where $p(\mathbf{x}|y)$ is the probability of sampling $\mathbf{x}$ conditioned on the class $y \in \{0, 1\}$ of the sample. Each $\mathbf{x}$ is sampled using a multivariate normal distribution. $\mathcal{S}^T(y)$ is the set of all possible indexes for class $y$ and dataset type $T$. $\mathcal{P}_\alpha$ is the probability of the $\alpha$-th $D$-dimensional multivariate normal distribution $\mathcal{N}(\mathbf{\mu_\alpha} , \mathbf{\Sigma_\alpha})$, which depends on the size of $\mathcal{S}^T(y)$. This formulation can easily be used to generate data with a generic number of clusters, classes, and dataset types. In particular, more classes and dataset types can be included by defining proper, additional sets of indexes $\mathcal{S}^T(y)$. For our experiments, we only considered two classes. 

\subsection{Data Generation Process}
\label{sec:data_gen}

\paragraph{Labels:}
The first step in generating the data was to create a vector $y$ of classes $0$ or $1$ of size equal to the desired number of training samples $N$. This was done by sampling $N$ values from a uniform distribution $\mathcal{U}(0,1)$ and reassigning them to values $0$ or $1$ depending on whether they were $ \le 0.5$ or $ > 0.5$, respectively. The resulting vector $\mathbf{y}$ is an array of values $0$ or $1$ in balanced proportions and it corresponds to the labels of the training samples in our dataset. 

\paragraph{Linear dataset:}
For each observation $\mathbf{x_i}$, $i=1,\dots,N$, of the linear dataset, the set of indexes $\mathcal{S}^L(y)$ has only one element for each class. This means that, for each class, the input points can be sampled by one multivariate normal:

\begin{subequations}
\label{eq:linear_mixture_model}
\begin{gather}
\mathcal{S}^L(y=0) = \{\alpha_0\} \; \rightarrow \; \mathbf{\mu}_{\alpha_0} = 0 \cdot \mathbbm{1}^D
\\
\mathcal{S}^L(y=1) = \{\alpha_1\} \; \rightarrow \; \mathbf{\mu}_{\alpha_1} = \nu \cdot \mathbbm{1}^D
\\
\mathbf{\Sigma}_{\alpha_0} = \mathbf{\Sigma}_{\alpha_1} = I_D.
\end{gather}
\end{subequations}

$\mathbbm{1}^D$ is a $D$-dimensional vector of all ones that can be multiplied by a scalar, meaning that all its elements get multiplied by that scalar. $I_D$ is the $D \times D$ identity matrix, and its elements can be multiplied by a scalar number as well\footnote{This notation is used consistently in this section, to indicate the means and covariances of the normal distributions.}. The distance between the two clusters, which makes the two classes more or less discernible, can be changed by simply increasing or decreasing the value of the $\nu$ coefficient. Performing PCA on the linear dataset and plotting the first two principal components yields the clusters shown in \cref{fig:linear_ds}. In those plots, it is possible to observe a clear distinction between the clusters belonging to the two classes. We can also see the change in distance between the clusters as $\nu$ is changed.

\paragraph{XOR dataset:}
For the second dataset, our goal was to make the data appear as four separate clusters placed like the visual representation of the XOR logical operator. In this case, the sets of indexes corresponding to class $y=0$ and $y=1$ that have two elements each. Thus, the data points of each class can be sampled from two different distributions with equal probabilities. For class $y=0$, the samples are generated like the linear dataset described in \cref{eq:linear_mixture_model}. For the other class, its main feature is that the mean vectors of the multivariate normal distributions consist of two $\frac{D}{2}$-dimensional halves with different values:

\begin{subequations}
\label{eq:xor_mixture_model}
\begin{gather}
\mathcal{S}^X(y=0) = \{\alpha_0^A, \alpha_0^B \} \; \rightarrow 
    \begin{cases}
     \mathbf{\mu}_{\alpha_0^A} = 0 \cdot \mathbbm{1}^D \\
     \mathbf{\mu}_{\alpha_0^B} = \nu \cdot \mathbbm{1}^D
    \end{cases}
\\
\mathcal{S}^X(y=1) = \{\alpha_1^A, \alpha_1^B \} \; \rightarrow 
    \begin{cases}
     \mathbf{\mu}_{\alpha_1^A} = [0 \cdot \mathbbm{1}^\frac{D}{2} \,, \, \nu \cdot \mathbbm{1}^\frac{D}{2}] \\
     \mathbf{\mu}_{\alpha_0^B} = [\nu \cdot \mathbbm{1}^\frac{D}{2} \,, \, 0 \cdot \mathbbm{1}^\frac{D}{2}]  
    \end{cases}
\\
\mathbf{\Sigma}_{\alpha_0^A} = \mathbf{\Sigma}_{\alpha_0^B} = \mathbf{\Sigma}_{\alpha_1^A} = \mathbf{\Sigma}_{\alpha_1^B} = I_D.
\end{gather}
\end{subequations}

The PC plots of this dataset are shown in \cref{fig:XOR_ds}, where we can see the positioning of the four clusters in a cross-like layout.

\section{Pruning and rewinding} \label{app:IMP}
In our experiments, training is always followed by a series of pruning iterations that were performed according to the IMP \citep{han2015learning, frankle2018lottery} technique, which is described below:

\begin{enumerate}
    \item Initialise the model with weight $\mathcal{W}_0$, obtaining a model function $f(x \, ; \, \mathcal{W}_0)$.
    \item Train the model for the desired number of epochs $j$, reaching a set of weights $\mathcal{W}_j$.
    \item Sort all the weights in $\mathcal{W}_j$ by their absolute value and select the lowest $p \%$, where $p$ is a number between 0 and 100.
    \item Prune the selected weights by applying a mask to the original model, obtaining $f(x \, ; \, \text{mask} \odot \mathcal{W}_j)$. 
    \item Reset the remaining parameters to their first initialisation values, obtaining $f(x \, ; \, \text{mask} \odot \mathcal{W}_0)$. 
    \item Iterate the process $n$ times, pruning $p\%$ of the remaining connections at each iteration.
\end{enumerate}

In certain cases, the larger models need to be pruned using the lottery ticket rewinding technique \citep{frankle2019stabilizing}. Rewinding simply consists of modifying only one step of the procedure described above:

\begin{enumerate}
\item[5] Reset the remaining parameters to the values at iteration $k \ll j$ of the training loop, obtaining $f(x \, ; \, \text{mask} \odot \mathcal{W}_k)$
\end{enumerate}

where $k$ is a hyperparameter representing the number of the rewinding iterations. This technique is also conveniently included in the \texttt{OpenLTH}\footnote{\url{https://github.com/facebookresearch/open_lth}} library.

\section{Model hyperparameters}
\label{app:hyperparameters}
\subsection{Mixture classification}
We used two-layer fully connected networks without bias for our experiments on the two mixture classification datasets. Models with $P \in \{1, 3, 5, 8, 10, 15, 20, 25, 50, 100, 200, 500, 1000, 10000\}$ neurons in the hidden layer were used to produce the double descent curve. These models were subsequently pruned using IMP to produce the sparse double descent curves. Network weights are initialised using the Kaiming uniform distribution. The activation function chosen for this model is the Rectified Linear Unit (ReLU). Stochastic gradient descent with a learning rate of 0.1 was used as the optimiser. All models were trained for \num{1000} epochs using a batch size of \num{1024}. All models were sufficiently pruned to observe the sparse double descent curves. All experiments were replicated five times.

\paragraph{Linear datasets:}
We varied the distance between cluster means $\nu \in \{0.1, 0.2, 0.3, 0.4, 0.5, 0.6, 0.7, 0.8, 0.9, 1.0\}$ to see how it impacts the test error and the effective number of parameters. 5\% random label noise in the training set was used to observe the two double descent curves.

\paragraph{XOR datasets:}
We varied the distance between cluster means $\nu \in \{0.6, 0.7, 0.8, 0.9, 1.0, 1.1, 1.2, 1.3\}$ to see how it impacts the test error and the effective number of parameters. Higher values of $\nu$ were needed to ensure sufficient distance between the clusters. Higher label noise of 25\% was needed to consistently observe the two double descent curves in the XOR datasets.

\subsection{MNIST}
The three-layer fully connected architecture used for MNIST is an extension of the standard model used for MNIST in the pruning literature. We varied the number of neurons $P \in \{3, 5, 10, 25, 50, 100, 300, 500, 1000, 5000, 10000\}$ in the first hidden layer. The size of the second hidden layer was kept fixed at 100. For the two-layer FCN, $P \in \{5, 10, 50, 100, 300, 500, 1000, 2000, 5000, 10000\}$ neurons were used in the hidden layer. For the fully connected networks, lottery ticket rewinding was used for networks with $P \geq 1000$. For ResNet-6, we varied the width $W \in \{1, 2, 5, 8, 11, 15, 20, 40, 80, 120\}$ of the convolutional filters to obtain networks of different sizes. Further details can be found in \cref{tab:model_details}.

\subsection{Fashion-MNIST}
We also performed a small set of experiments on the Fashion-MNIST dataset. We only performed a limited number of experiments focused on finding the effective number of parameters using overparameterised models. Since reproducing the double descent curve was not the target, we only used three-layer FCNs with $P \in \{500, 1000\}$ neurons in the first hidden layer. The size of the second hidden layer was kept fixed at 100. Note that more epochs (320) are needed to train a three-layer FCN on Fashion-MNIST.

\subsection{CIFAR-10}
We primarily considered ResNet-18 for CIFAR-10, where the width $W \in \{2, 5, 8, 11, 15, 20, 40, 60, 80, 100, 120, 150\}$ of the convolutional filters was varied to obtain networks of different sizes. Based on previous observations \citet{frankle2019stabilizing, he2022sparse}, we used 10 epochs of rewinding to consistently find lottery tickets in ResNet-18. To compare the effective number of parameters across different CNN architectures, we performed our analysis on DenseNet-121 and VGG-16. Further details can be found in \cref{tab:model_details}.

\begin{table}[!ht]
    \centering
    \caption{Neural network architectures used on real data. LR refers to the learning rate.}
    \label{tab:model_details}
    \begin{tabular}{l|lrrrrrr}
        Network & Dataset & Epochs & Batch size & Optimiser & Momentum & LR & Rewind Iter \\ \hline
        Two-layer FCN & MNIST & 120 & 128 & SGD & - & 0.1 & - \\
        Three-layer FCN & MNIST & 120 & 128 & SGD & - & 0.1 & - \\
        ResNet-6 & MNIST & 120 & 128 & SGD & 0.9 & 0.1 & - \\
        Three-layer FCN & Fashion MNIST & 320 & 128 & SGD & - & 0.1 & - \\
        ResNet-18 & CIFAR-10 & 160 & 128 & SGD & 0.9 & 0.1 & 10 epochs \\
        VGG-16 & CIFAR-10 & 160 & 128 & SGD & 0.9 & 0.1 & 10 epochs \\
        DenseNet-121 & CIFAR-10 & 160 & 128 & SGD & 0.9 & 0.1 & 10 epochs \\
    \end{tabular}
\end{table}

\section{Additional Results}
\begin{figure}[!ht]
\begin{subfigure}{.48\textwidth}
    \centering
    \includegraphics[width=\linewidth]{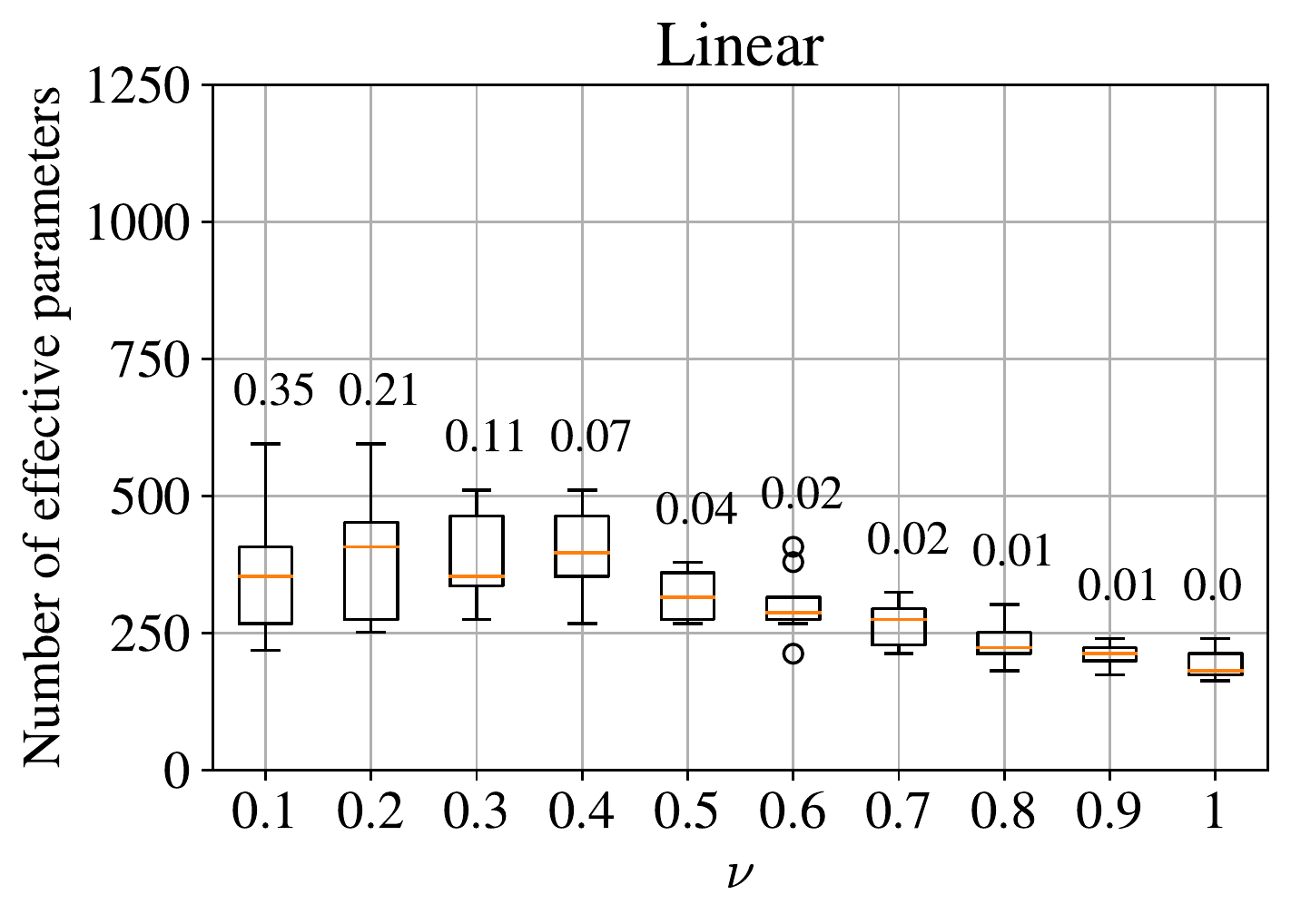}
    \label{fig:linear_neff}
\end{subfigure} \hfill
\begin{subfigure}{.48\textwidth}
    \centering
    \includegraphics[width=\linewidth]{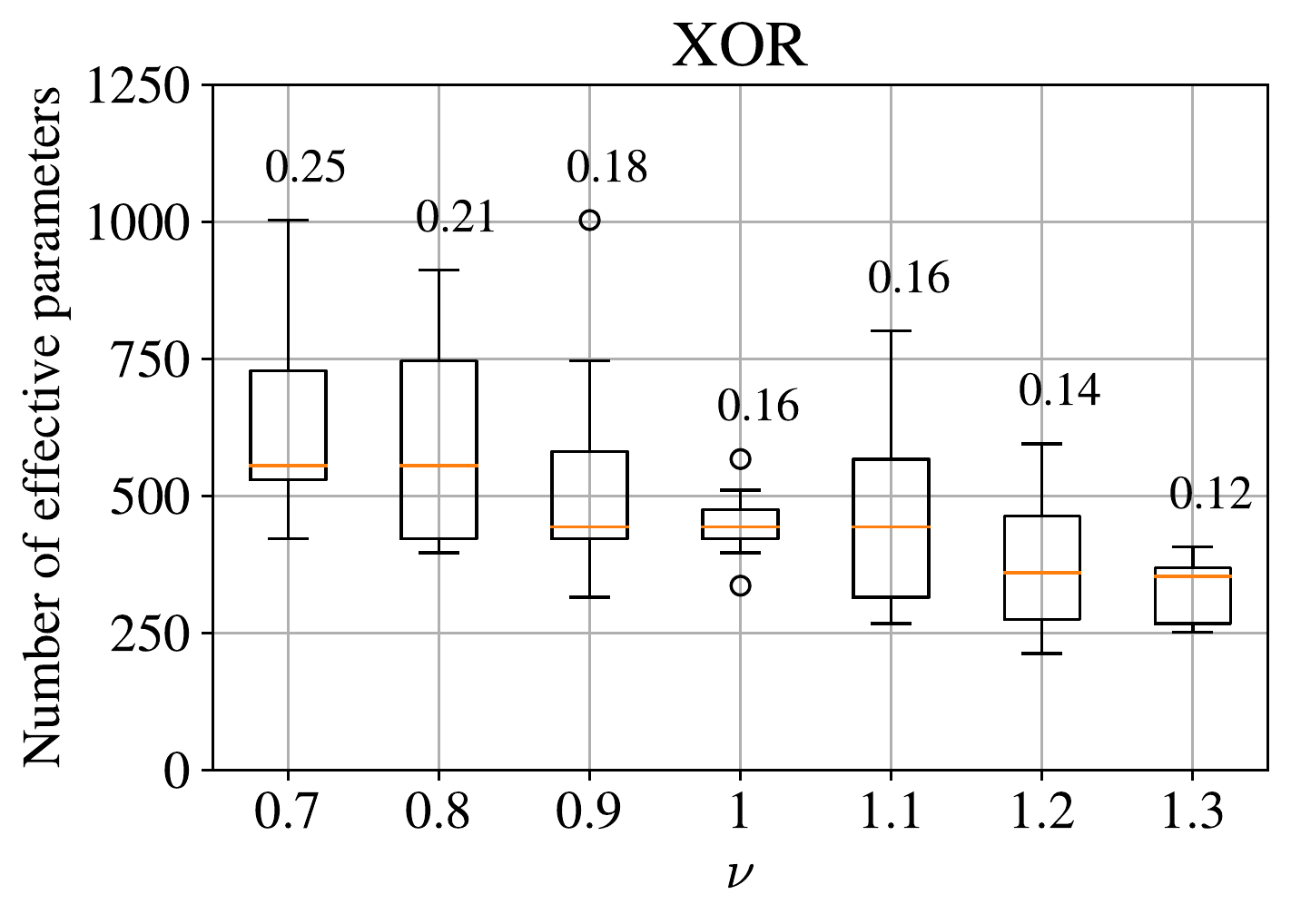}
    \caption{XOR datasets.}
    \label{fig:xor_neff}
\end{subfigure}
    \caption{Distribution of the effective number of parameters of the best pruned models (y-axis) as the distance between the clusters $\nu$ (x-axis) is varied for the linear and xor datasets. Only pruned models originating from overparameterised full models are considered. The numbers above the boxes report the test error of the model with median effective number of parameters for each $\nu$.}
    \label{fig:neff_comparison}
\end{figure}

\Cref{fig:neff_comparison} shows how the effective number of parameters and test error of the best pruned models vary with $\nu$ for the linear and XOR datasets. As expected, the test error decreases with the distance between cluster means $\nu$. For networks trained on the linear dataset, we find that the effective number of parameters, i.e.~the number of parameters in the best pruned models, ranges from $\sim \num{200}$ to $\sim \num{500}$ for different starting models and for different values of $\nu$ (see \cref{fig:linear_neff} for the full distribution). We find the same behaviour for the XOR dataset, \cref{fig:XOR_ds}, but with a higher number of parameters (between $\num{250}$ and $\num{1000}$) than for the linear dataset with the same $\nu$.

\begin{figure}[!ht]
    \centering
    \includegraphics[width=.48\linewidth]{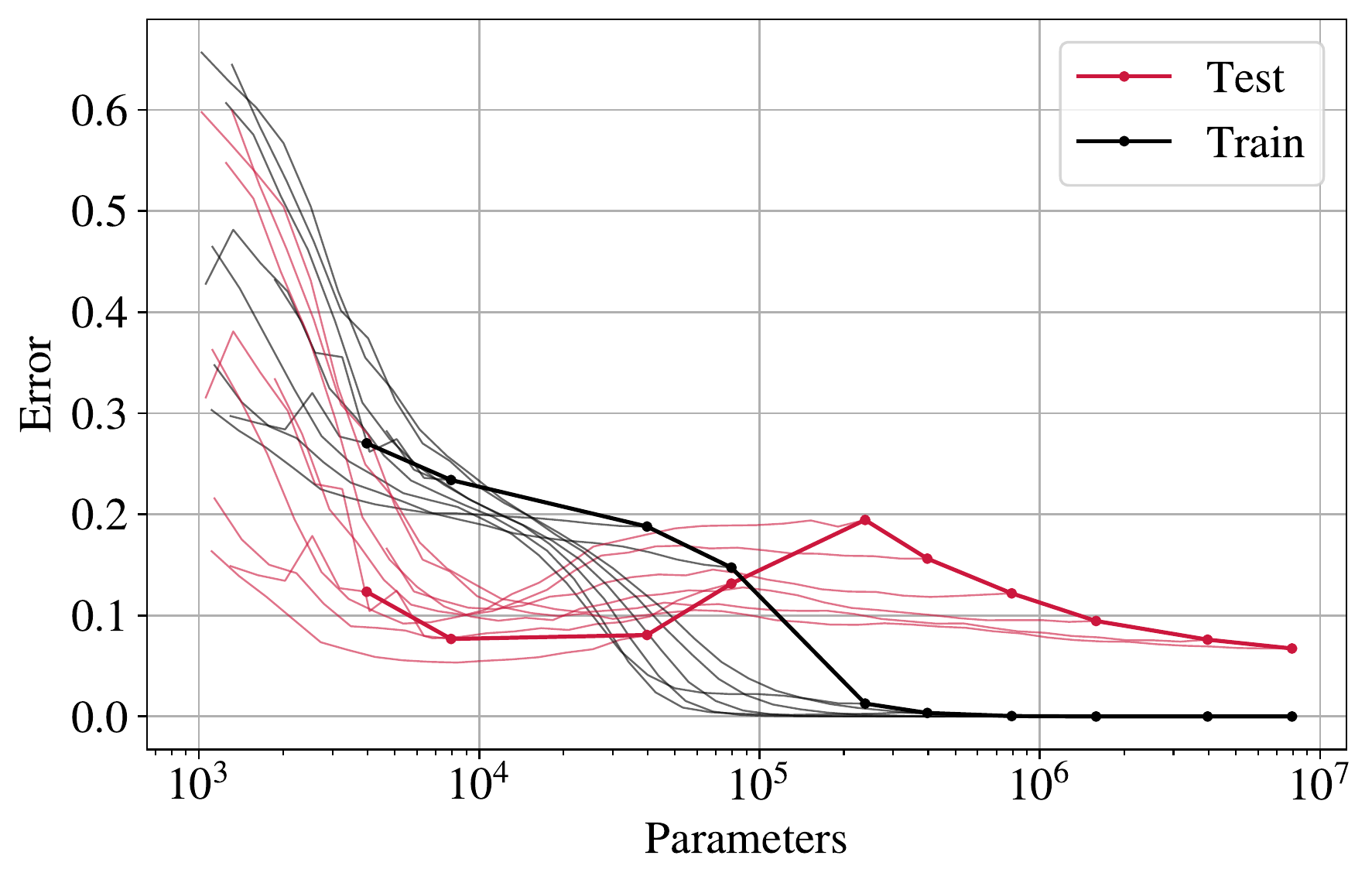} \hfill
    \includegraphics[width=.48\linewidth]{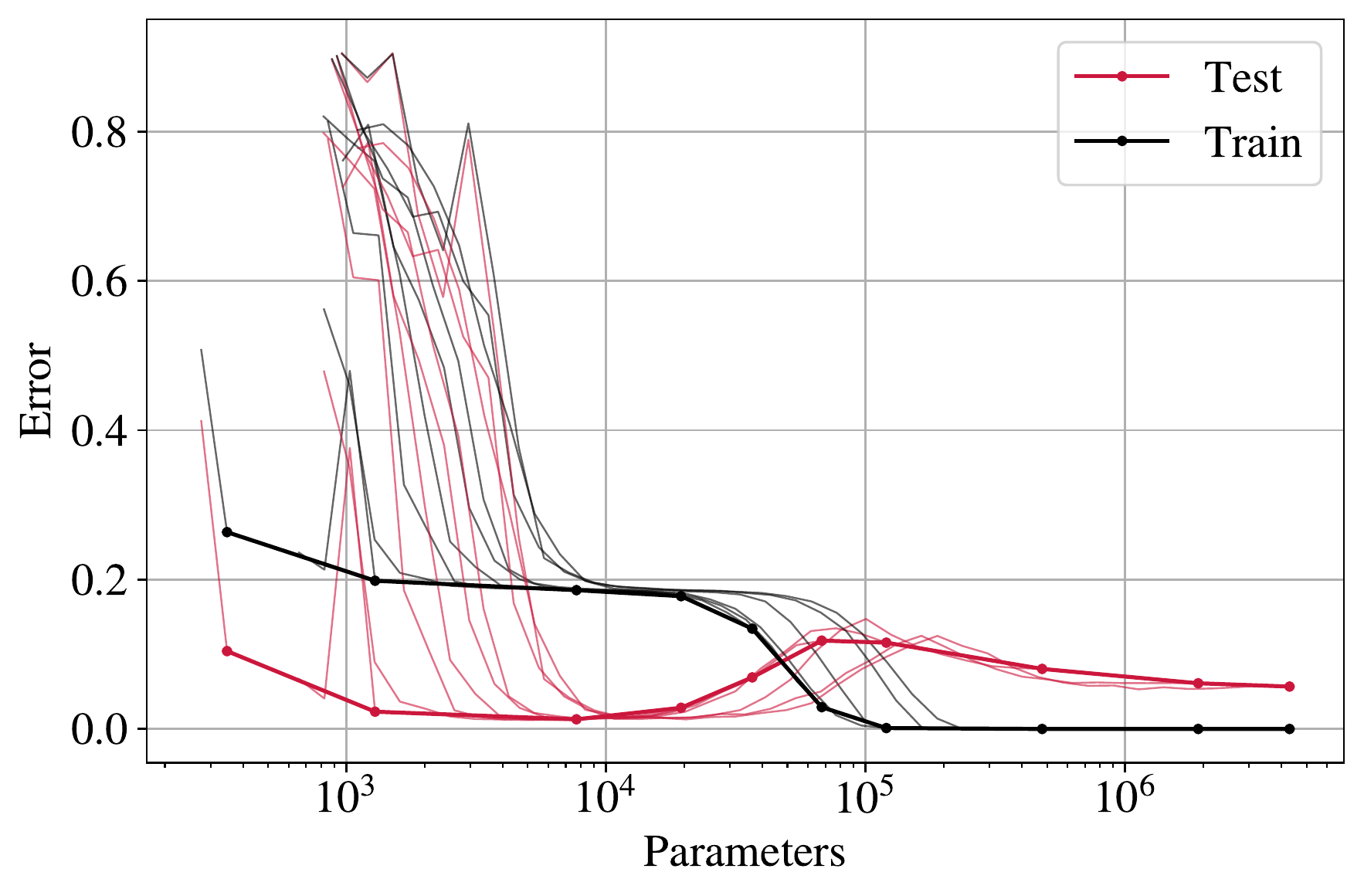}
    \caption{Pruning models along the double descent curve (dark red) show that sparse double descent curves (light red) from different models coincide at the minima. Results are shown for two-layer FCNs (left) and ResNet-6 (right) on MNIST with 20\% label noise averaged over three replicates.}
    \label{fig:mnist_app}
\end{figure}

\begin{figure}[!ht]
    \centering
    \includegraphics[width=.48\linewidth]{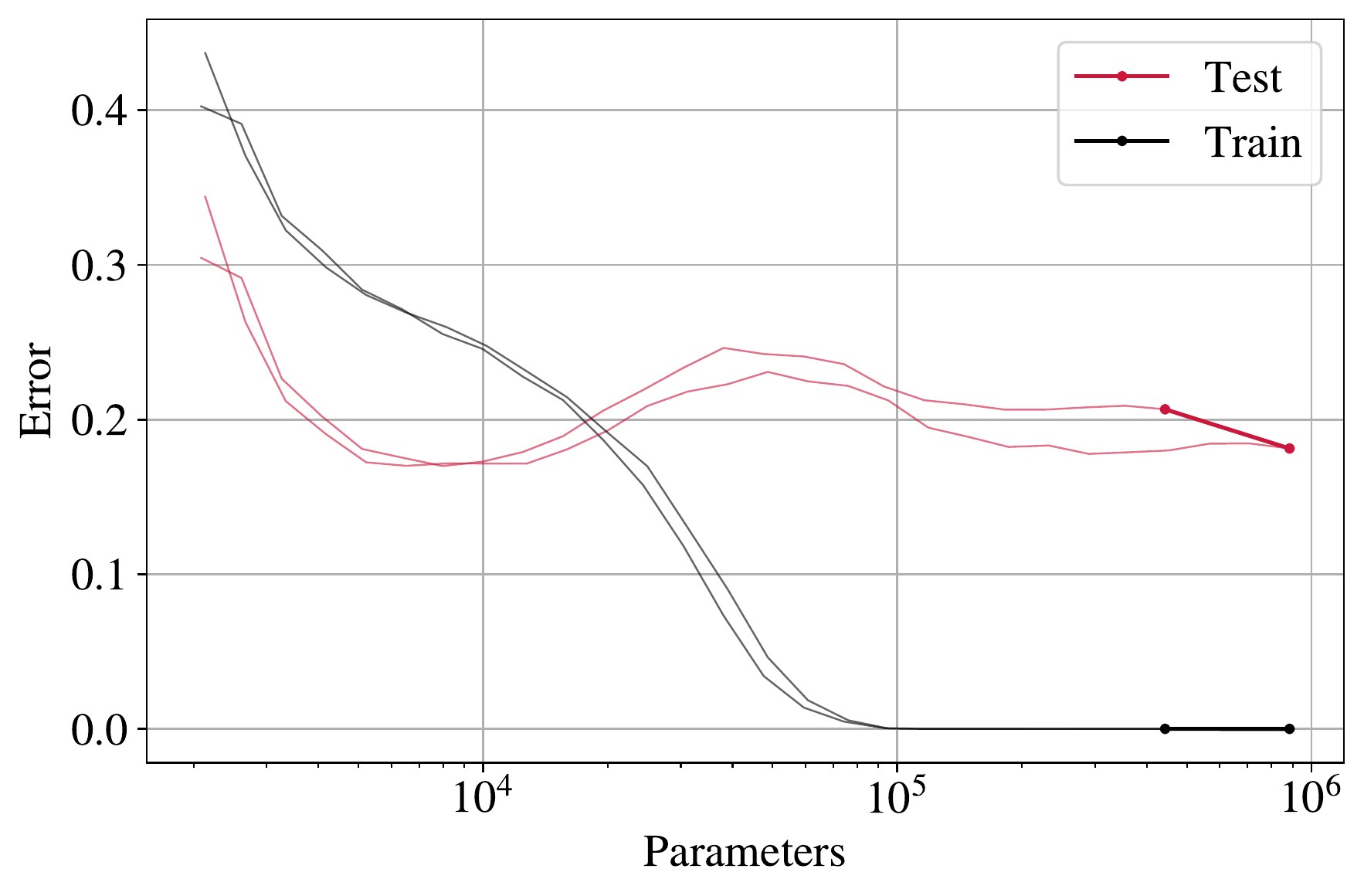}
    \caption{Estimating the effective number of parameters for Fashion-MNIST: Results are shown for three-layer FCNs on Fashion-MNIST with 20\% label noise averaged over three replicates. Compared to MNIST, the test error for the best pruned models is higher while the effective number of parameters is approximately \num{10000}.}
    \label{fig:fmnist}
\end{figure}

Double descent and sparse double descent curves obtained for MNIST on two-layer FCN and ResNet-6 can be seen in \cref{fig:mnist_app}. \Cref{tab:summary_mnist} shows the number of parameters and test errors for the full/unpruned models and the corresponding best pruned models. Sparse double descent curves for two different models on Fashion-MNIST in \cref{fig:fmnist} show that, compared to MNIST, the test error for the best pruned models is higher while the effective number of parameters is approximately \num{10000}.

\begin{table}[!ht]
    \centering
    \caption{Number of parameters and test error for unpruned and best pruned models for two architectures trained on MNIST: two-layer FCNs, and ResNet-6. Average values over 3 replicates are reported. We observe that a $\num{200}\times$ increase for the full models results in only a $\sim\num{3.5}\times$ increase in the number of parameters for the best pruned models. Notice also that the error achieved by pruned models appears insensitive to the error rate of the original full model, i.e. even models with poor generalisation can be rescued by pruning.}
    \label{tab:summary_mnist}
    \begin{tabular}{rr|rr||rr|rr}
       \multicolumn{4}{c||}{2 layer FC} & \multicolumn{4}{c}{ResNet-6} \\ \hline
       \multicolumn{2}{c|}{Parameters} & \multicolumn{2}{c||}{Test error} & \multicolumn{2}{c|}{Parameters} & \multicolumn{2}{c}{Test error} \\ \hline
       Full & Pruned & Full & Pruned & Full & Pruned & Full & Pruned \\ \hline
        \num{39700} & \num{8320} & \num{0.081} & \num{0.053} & \num{19464} & \num{5098} & \num{0.028} & \num{0.012} \\
        \num{79400} & \num{6814} & \num{0.131} & \num{0.077} & \num{36597} & \num{7669} & \num{0.069} & \num{0.012} \\
        \num{238200} & \num{5358} & \num{0.194} & \num{0.092} & \num{67785} & \num{7272} & \num{0.118} & \num{0.014} \\
        \num{794000} & \num{11436} & \num{0.122} & \num{0.106} & \num{120180} & \num{8252} & \num{0.116} & \num{0.013} \\ 
        \num{1588000} & \num{11710} & \num{0.094} & \num{0.095} & \num{478760} & \num{13470} & \num{0.080} & \num{0.013} \\
        \num{3970000} & \num{23428} & \num{0.076} & \num{0.103} & \num{1911120} & \num{17620} & \num{0.061} & \num{0.014} \\
        \num{7940000} & \num{29990} & \num{0.067} & \num{0.097} & \num{4297080} & \num{20286} & \num{0.057} & \num{0.013} \\
    \end{tabular}
\end{table}

\begin{table}[!ht]
    \caption{Effective number of parameters and test error of best pruned models on subsets of CIFAR-10 dataset.}
    \label{tab:cifar_sub}
    \centering
    \begin{tabular}{c|rr}
       Classes & Params & Error \\ \hline
        5 & \num{16312} & \num{0.102} \\
        6 & \num{20392} & \num{0.118} \\
        7 & \num{31869} & \num{0.122} \\
        8 & \num{31871} & \num{0.125} \\
        9 & \num{31872} & \num{0.126} \\
        10 & \num{39843} & \num{0.124} \\
    \end{tabular}
\end{table}

\begin{figure*}[!ht]
    \centering
    \begin{subfigure}{.49\linewidth}
        \includegraphics[width=\textwidth]{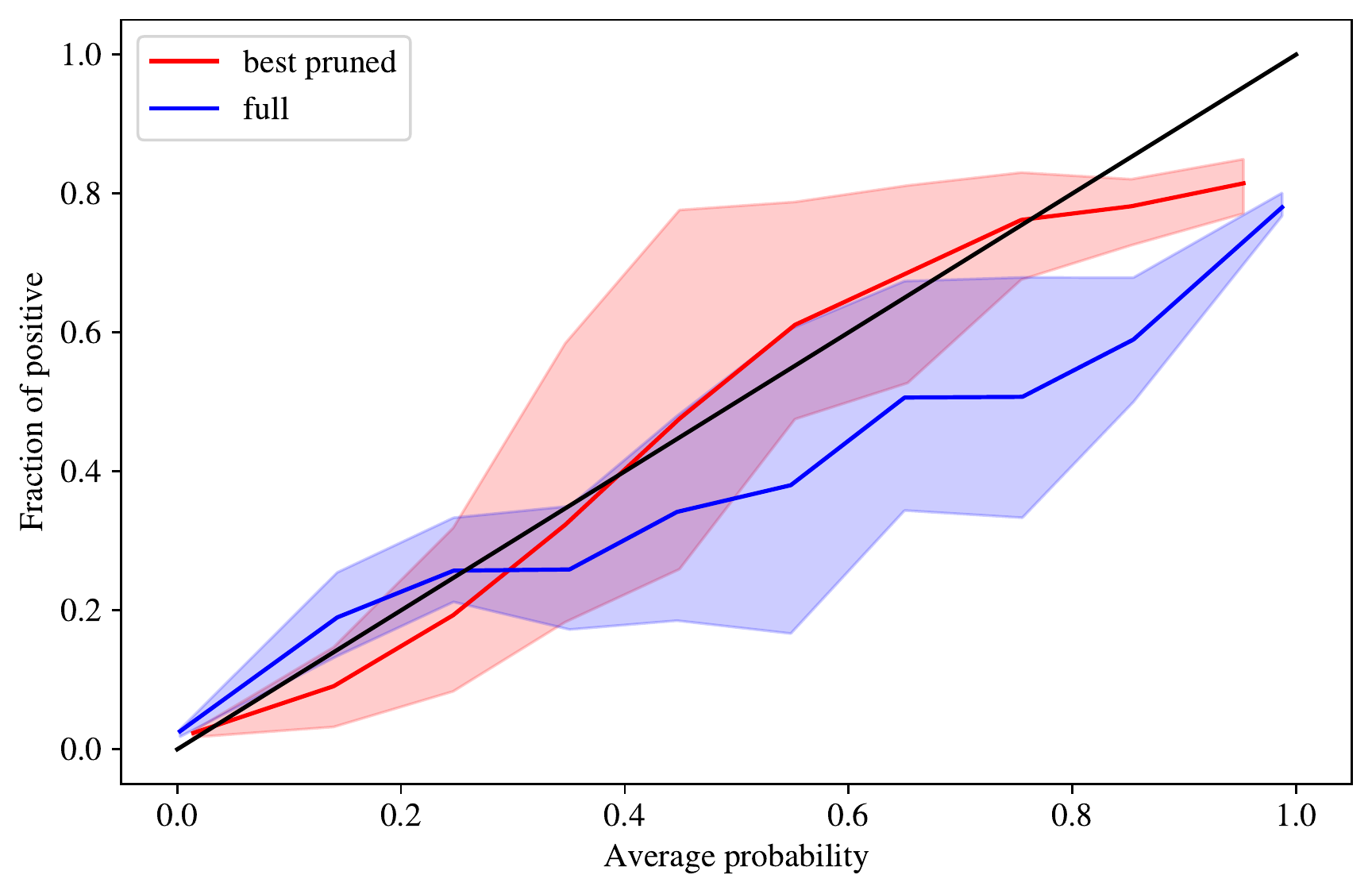}
        \caption{MNIST with two-layer FCN. ECE for pruned and full models are $0.075$ and $0.207$, respectively.} \label{fig:calib_mnist_noise}
    \end{subfigure}\hfill
    \begin{subfigure}{.49\linewidth}
        \includegraphics[width=\textwidth]{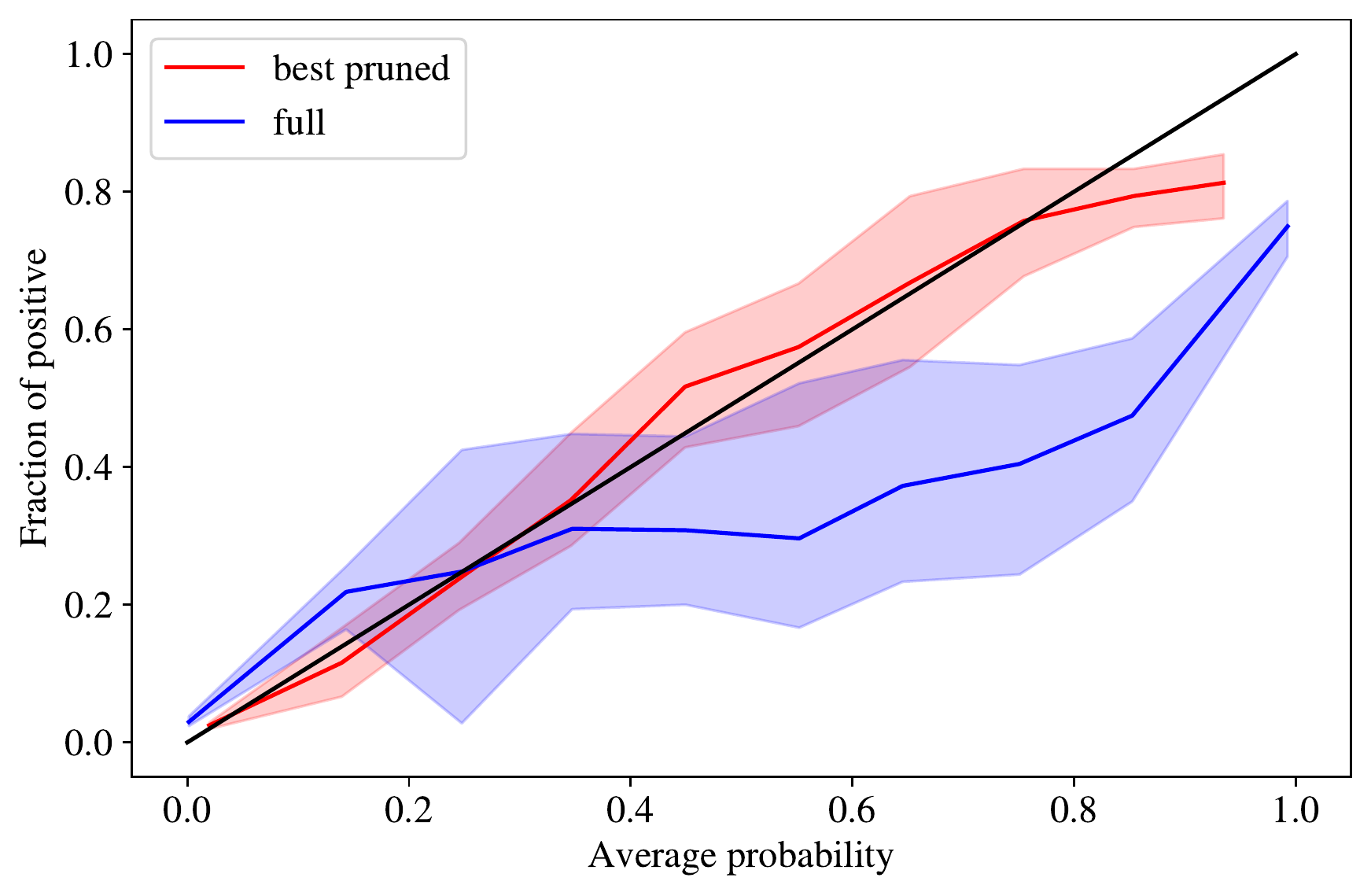}
        \caption{CIFAR-10 with ResNet-18. ECE for pruned and full models are $0.053$ and $0.254$, respectively.} \label{fig:calib_cifar_noise}
    \end{subfigure}
    \caption{Class-averaged calibration curves for the best pruned and full models on (a) MNIST and (b) CIFAR10 datasets with noise added to the labels in the test set show that the pruned models are well-calibrated to noisy data while the full models are overconfident. The highlighted areas signify deviation between classes.}
    \label{fig:calib_real_noise}
\end{figure*}

Finally, the calibration curves when label noise is added to the test set for MNIST and CIFAR-10 can be seen in \cref{fig:calib_real_noise}. The class-averaged calibration curves show that the pruned models are well-calibrated to noisy data while the full models are overconfident.

\end{document}